\documentclass[10pt,twocolumn,letterpaper]{article}

\usepackage{iccv}
\usepackage{lipsum}
\usepackage{times}
\usepackage{epsfig}
\usepackage{graphicx}
\usepackage{amsmath}
\usepackage{amssymb}
\usepackage{cite}
\usepackage{multicol}
\usepackage{wrapfig}
\usepackage[utf8]{inputenc} 
\usepackage[T1]{fontenc}    
\usepackage{url}            
\usepackage{booktabs}       
\usepackage{amsfonts}       
\usepackage{nicefrac}       
\usepackage{microtype}      

\usepackage{epsfig}
\usepackage{graphicx}
\usepackage{amsmath}
\usepackage{amsthm}
\usepackage{amssymb}
\usepackage{dsfont}
\usepackage{subfigure}
\usepackage{color}

\usepackage[font=small,skip=5pt]{caption}

\newtheorem{thm}{Theorem}

\usepackage[pagebackref=true,breaklinks=true,letterpaper=true,colorlinks,bookmarks=false]{hyperref}

\iccvfinalcopy 

\def\iccvPaperID{4171} 

\ificcvfinal\fi
\begin{document}

\title{\vspace{-10mm}MONET: Multiview Semi-supervised Keypoint Detection\\via Epipolar Divergence}

\author{Yuan Yao\\
University of Minnesota\\
{\tt\small yaoxx340@umn.edu}
\and
Yasamin Jafarian\\
University of Minnesota\\
{\tt\small yasamin@umn.edu}
\and
Hyun Soo Park\\
University of Minnesota\\
{\tt\small hspark@umn.edu}
}


\twocolumn[{%
\maketitle
\begin{center}
\vspace{-7mm}
		\centering
	\includegraphics[width=\textwidth]{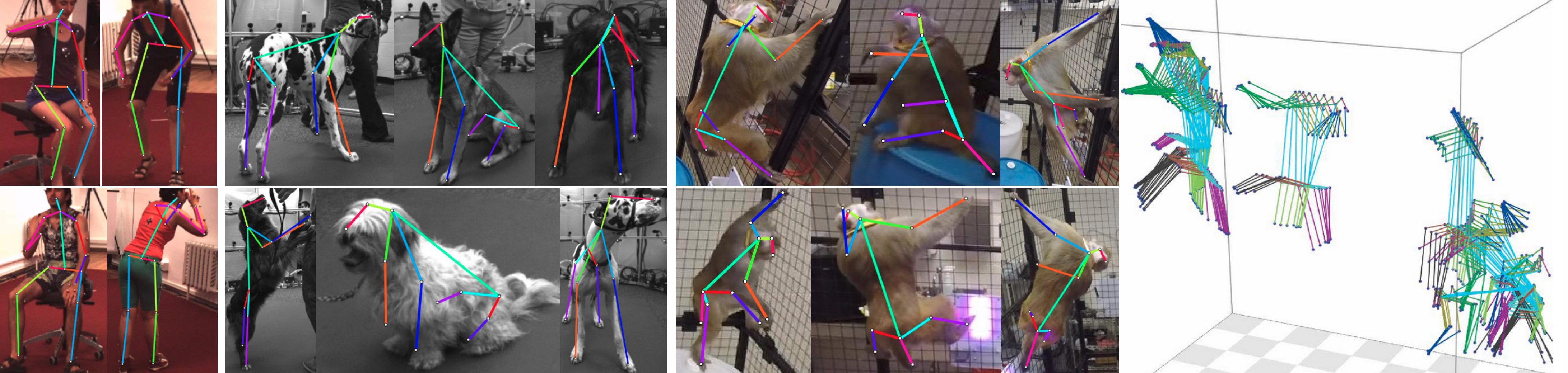}
	\captionof{figure}{This paper presents MONET-an semi-supervised learning for keypoint detection, which is able to localize customized keypoints of diverse species, e.g., humans, dogs, and monkeys with very limited number of labeled data without a pre-trained model. The right most figure illustrates 3D reconstruction of monkey movement using our pose detection.}
	\label{fig:teaser}
	\vspace{7mm}
\end{center}	
\label{fig:teaser_main}
}]

\begin{abstract}
   This paper presents MONET---an end-to-end semi-supervised learning framework for a keypoint detector using multiview image streams. In particular, we consider general subjects such as non-human species where attaining a large scale annotated dataset is challenging. While multiview geometry can be used to self-supervise the unlabeled data, integrating the geometry into learning a keypoint detector is challenging due to representation mismatch. We address this mismatch by formulating a new differentiable representation of the epipolar constraint called epipolar divergence---a generalized distance from the epipolar lines to the corresponding keypoint distribution. Epipolar divergence characterizes when two view keypoint distributions produce zero reprojection error. We design a twin network that minimizes the epipolar divergence through stereo rectification that can significantly alleviate computational complexity and sampling aliasing in training. We demonstrate that our framework can localize customized keypoints of diverse species, e.g., humans, dogs, and monkeys.
\end{abstract}

\section{Introduction}

Human pose detection has advanced significantly over the last few
years~\cite{cao2017realtime,wei2016cpm,newell:2016,toshev:2014}, driven in large part to new approaches based on deep learning. But these techniques require large amounts of labeled
training data. For this reason, pose detection is almost always
demonstrated on humans, for which large-scale datasets are available
(e.g., MS COCO~\cite{lin:2014} and MPII~\cite{Andriluka:2014}). What about pose detectors for other animals, such as
monkeys, mice, and dogs? Such algorithms could have enormous
scientific impact~\cite{Mathisetal2018}, but obtaining large-scale labeled training data
would be a substantial challenge: each individual species may need its
own dataset, some species have large intra-class variations, and
domain experts may be needed to perform the labeling accurately.
Moreover, while there is significant commercial interest in human pose
recognition, there may be little incentive for companies and research
labs to invest in collecting large-scale datasets for other species.

This paper addresses this annotation challenge by leveraging {\em multiview image streams}. Our insight is that the manual effort of annotation can be significantly reduced by using the redundant visual information embedded in the multiview imagery, allowing cross-view self-supervision: one image can provide a supervisionary signal to another image through epipolar geometry without 3D reconstruction. To this end, we design a novel end-to-end semi-supervised framework to utilize a large set of unlabeled multiview images using cross-view supervision.

\begin{figure*}[t]
\begin{center}
\subfigure[Representation mismatch]{\label{Fig:rep1}\includegraphics[height=0.16\textwidth]{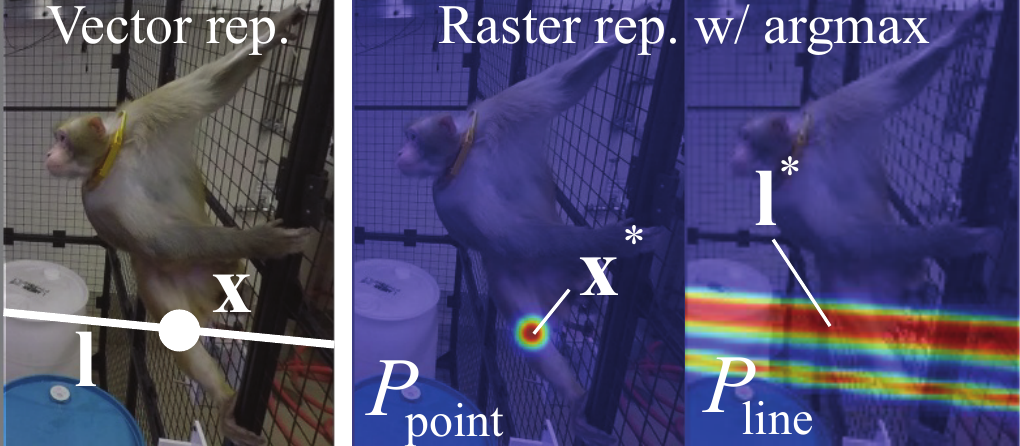}}~~
\subfigure[Triangulation]{\label{Fig:triangulation}\includegraphics[height=0.16\textwidth]{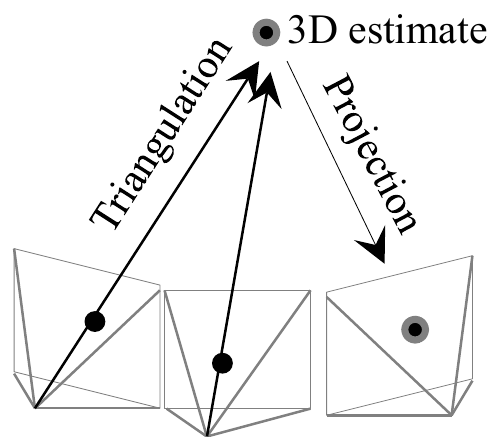}}~~
\subfigure[Depth prediction]{\label{Fig:depth_prediction}\includegraphics[height=0.16\textwidth]{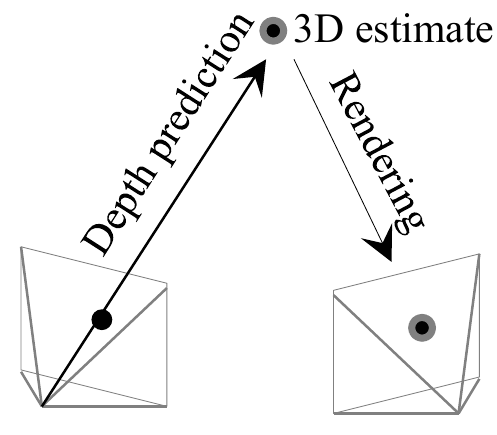}}~~
\subfigure[Epipolar line]{\label{Fig:transfer_epiplarplane}\includegraphics[height=0.14\textwidth]{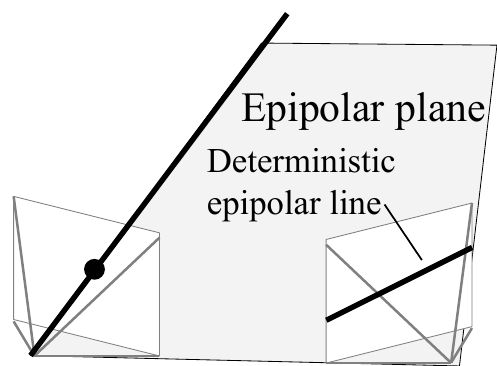}}
\end{center}
\vspace{-3mm}
   \caption{(a) Vector and raster representations describe the epipolar geometry. Note that the raster representation requires a non-differentiable argmax operation to compute $\mathbf{x}^*$ and $\mathbf{l}^*$. (b-d) Various multiview supervision approaches. (b) Keypoint prediction from at least two images can be triangulated and projected to supervise another image. This involves a non-differentiable argmax and RANSAC process~\cite{simon:2017}. (c) A 3D point~\cite{Rhodin:2018cvpr}, mesh~\cite{Kanazawa:2018}, and voxel~\cite{NIPS2016_6206} can be predicted from a single view and projected to supervise another image. This requires an additional 3D prediction that fundamentally bounds the supervision accuracy. (d) Our approach precisely transfers a keypoint detection in one image to another image through the epipolar plane for the cross-supervision, and does not require 3D reconstruction.  }
\label{Fig:ge}
\vspace{-4mm}
\end{figure*}

The key challenge of integrating the epipolar geometry for building a strong keypoint (pose) detector lies in a representational mismatch: the geometric quantities such as points, lines, and planes are represented as a {\em vectors}~\cite{hartley:2004} (Figure~\ref{Fig:rep1} left) while the {\em raster} representation via pixel response (heatmap~\cite{wei2016cpm,cao2017realtime,newell:2016}) has been shown strong performance on keypoint detection. For instance, applying the epipolar constraint~\cite{longuet-higgins:1981}---a point $\mathbf{x}\in\mathds{R}^2$ must lie in the corresponding epipolar line $\mathbf{l}\in\mathds{P}^2$---can be expressed as:
\begin{align}
    &(\widetilde{\mathbf{x}}^*)^\mathsf{T}\mathbf{l}^* = 0 
    &{\rm s.t.~~} 
    \mathbf{x}^* = \underset{\mathbf{x}}{\operatorname{argmax}}~P_{\rm p}(\mathbf{x})
    ,~~\mathbf{l}^* = \underset{\mathbf{l}}{\operatorname{argmax}} ~P_{\rm e}(\mathbf{l}), \nonumber 
\end{align}
where $\widetilde{\mathbf{x}}$ is the homogeneous representation of $\mathbf{x}$, and $P_{\rm p}$ and $P_{\rm e}$ are the distributions of keypoints and epipolar lines\footnote{See Section~\ref{Sec:divergence}, respectively, as shown in Figure~\ref{Fig:rep1} for computation of $P_{\rm e}$.}. Note that the raster representation involves non-differentiable {\em argmax} operations, which are not trainable. This challenge leads to offline  reconstruction~\cite{simon:2017,Vijayanarasimhan:2017,Byravan:2016}, data driven depth prediction~\cite{Rhodin:2018eccv,Rhodin:2018cvpr,Angjoo:2018,drcTulsiani17,zhou2017unsupervised}, or the usage of the soft-argmax operation~\cite{dong2018supervision}, which shows inferior performance (see Figure~\ref{Fig:hypothesis}).

In this paper, we formulate a new raster representation of the epipolar geometry that eliminates the argmax operations. We prove that the minimization of geometric error (i.e., $|\widetilde{\mathbf{x}}^\mathsf{T}\mathbf{l}|$) is equivalent to minimizing {\em epipolar divergence}---a generalized distance from the epipolar lines to the corresponding keypoint distribution. 
With this measure, we design a new end-to-end semi-supervised network called MONET (Multiview Optical Supervision Network). The network efficiently leverages the unlabeled multiview image streams with limited numbers of manual annotations ($<1\%$). We integrate this raster formulation into the network by incorporating it with stereo rectification, which reduces the computational complexity and sampling artifacts while training the network.

The key features of MONET include that (1) it does not require offline triangulation that involves non-differentiable argmax and RANSAC operations~\cite{simon:2017} (Figure~\ref{Fig:triangulation}); (2) it does not require 3D prediction~\cite{Rhodin:2018cvpr,Rhodin:2018eccv,NIPS2016_6206} (Figure~\ref{Fig:depth_prediction}), i.e., it deterministically transfers keypoint detections in one image to the other via epipolar geometry (Figure~\ref{Fig:transfer_epiplarplane})\footnote{This is analogous to the fundamental matrix computation without 3D estimation~\cite{longuet-higgins:1981,hartley:2004}.}; (3) it is compatible with any keypoint detector design including CPM~\cite{wei2016cpm} and Hourglass~\cite{newell:2016} which localizes keypoints through a raster representation; and (4) it can apply to general multi-camera systems (e.g., different multi-camera rigs, number of cameras, and intrinsic parameters).

 The main contributions of this paper include:
(1) introducing a novel measure called the epipolar divergence, which
measures the geometric consistency between two view keypoint
distributions; (2) a network called MONET that efficiently minimizes
the epipolar divergence via stereo rectification of keypoint
distributions; (3) a technique for large-scale spatiotemporal data
augmentation using 3D reconstruction of keypoint trajectories; (4)
experimental results that demonstrate that MONET is flexible enough to
detect keypoints in various subjects (humans, dogs, and monkeys) in
different camera rigs and to outperform existing baselines in terms of
localization accuracy and precision (re-projection error). 


\section{Related Work}
The physical and social behaviors of non-human species such as rhesus macaque monkeys 
have been widely used as a window to study human activities in neuroscience and psychology. While measuring their subtle behaviors in 
the form of 3D anatomic landmarks is key, implementing marker-based 3D tracking systems is challenging due to the animal's sensitivity 
to reflective markers and occlusion by fur, which limits 
its applications 
to restricted body motions (e.g., body tied to a chair)~\cite{anderson:2008}. Vision-based marker-less motion capture is a viable solution to measure their free ranging behaviors~\cite{foster:2014,Sellers:2014,Nakamura:2016}.

In general, the number of 3D pose configurations of a deformable articulated body is exponential with respect to the number of joints. 
The 2D projections of the 3D body introduces substantial variability
in illumination, appearance, and occlusion, which makes
 pose estimation challenging. But the space of possible pose
configurations has structure that
can be captured by efficient spatial representations such
 as pictorial structures~\cite{Andriluka:2010,Andriluka:2009,Felzenszwalb:2005,Johnson:2010,Pishchulin1:2013,Pishchulin2:2013,Yang:2011}, hierarchical and
 non-tree models~\cite{Sun:2011,Tian:2012,Dantone:2013,Karlinsky:2012,Lan:2005,Sigal:2006,Wang:2008} and convolutional
 architectures~\cite{Shelhamer:2015,Pinheiro:2014,Carreira:2016,Chen:2014,Ouyang:2014,Pfister:2015,Tompson:2014,toshev:2014,krizhevsky:2012}, and inference
on these structures can be performed efficiently using clever
algorithms, e.g., dynamic programming, convex
 relaxation, and approximate algorithms.
Albeit efficient and accurate on canonical images, they exhibit inferior performance on images in the long-tail distribution, e.g., a pigeon pose of yoga. Fully supervised learning frameworks using millions of perceptrons in convolutional neural networks (CNNs)~\cite{cao2017realtime,wei2016cpm,newell:2016,toshev:2014} can address this long-tail distribution issue by leveraging a sheer amount of training data annotated by crowd workers~\cite{lin:2014,Andriluka:2014,Shotton:2011}. However, due to the number of parameters in a CNN, the trained model can be highly biased 
when the number of data samples is not sufficient ($<$1M).

Semi-supervised and weakly-supervised learning frameworks train CNN models with limited number of training data~\cite{SemiSupervised1,Zhou:2017,Belagiannis:2016,Tekin:2016,Iqbal:2017,Park:2017,Liu:2017,Lin:2017,Pavlakos:2017,Song:2017}.
For instance, temporal consistency derived by tracking during training can provide a supervisionary signal for body joint detection~\cite{Lin:2017}. Geometric (such as 3DPS model~\cite{Belagiannis:2016}) and spatial~\cite{Song:2017} relationship are another way to supervise body keypoint estimation.
Active learning that finds the most informative images to be annotated can alleviate the amount of labeling effort~\cite{Liu:2017}, and geometric~\cite{Pavlakos:2017} and temporal~\cite{Iqbal:2017} in 2D~\cite{TLD} and 3D~\cite{Yoon:2018,joo_cvpr_2014} consistency can also be used to augment annotation data. 

These approaches embed underlying spatial structures such as 3D skeletons and meshes that regularize the network weights. For instance, motion capture data 
can be used to jointly learn 2D and 3D keypoints~\cite{Zhou:2017}, and scanned human body models are used to validate 2D pose estimation via reprojection~\cite{Guler:2018,Joo:2018,Angjoo:2018,Yu:2018,Zuffi:2017}, e.g., by using a DoubleFusion system that can simultaneously reconstruct the inner body shape and pose. The outer surface geometry and motion in real-time by using a single depth camera~\cite{Yu:2018} and recovery human meshes that 
can reconstruct a full 3D mesh of human bodies from a single RGB camera by having 2D ground truth annotations~\cite{Angjoo:2018}. Graphical models can also be applied for animal shape reconstruction by learning a 3D model based on a small set of 3D scans of toy figurines in arbitrary poses and refining the model and initial registration of scans together, and then generalizing it by fitting the model to real images of animal species out of the training set~\cite{Zuffi:2017}.
Notably, a multi-camera system can be used to cross-view supervise multiview synchronized images using iterative process of 3D reconstruction and network training~\cite{simon:2017,Rhodin:2018cvpr}. 


Unlike existing methods, MONET does not rely on a spatial model. To our knowledge, this is the first paper that jointly reconstructs and trains a keypoint detector without iterative processes using epipolar geometry. We integrate reconstruction and learning through a new measure of keypoint distributions called epipolar divergence, which can apply to general subjects including non-human species where minimal manual annotations are available.

\begin{figure*}
\begin{center}
\subfigure[Geometric consistency via minimizing epipolar divergence ]{\label{Fig:geom1}\includegraphics[height=0.163\textwidth]{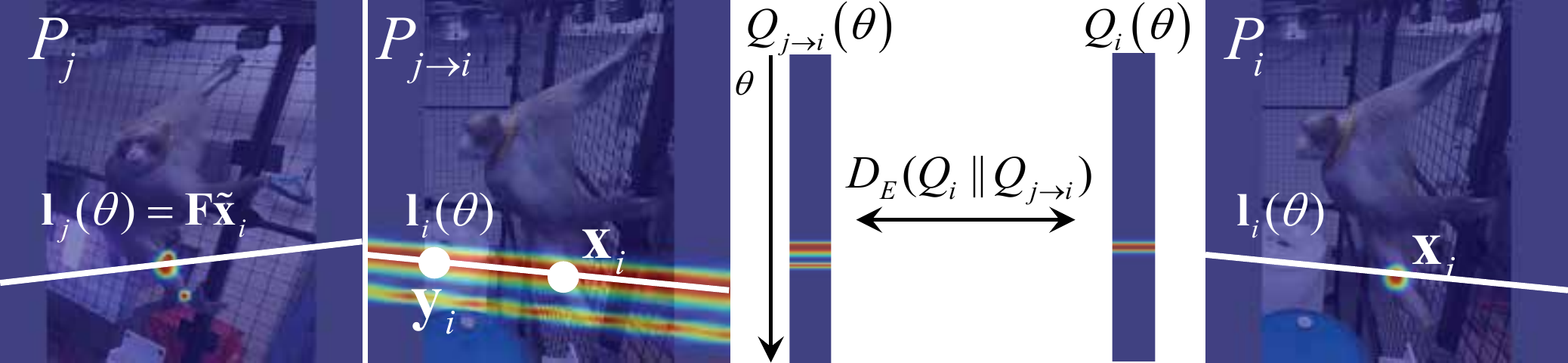}}~~~~~~~~
\subfigure[Epipolar plane parametrization]{\label{Fig:geom2}\includegraphics[height=0.163\textwidth]{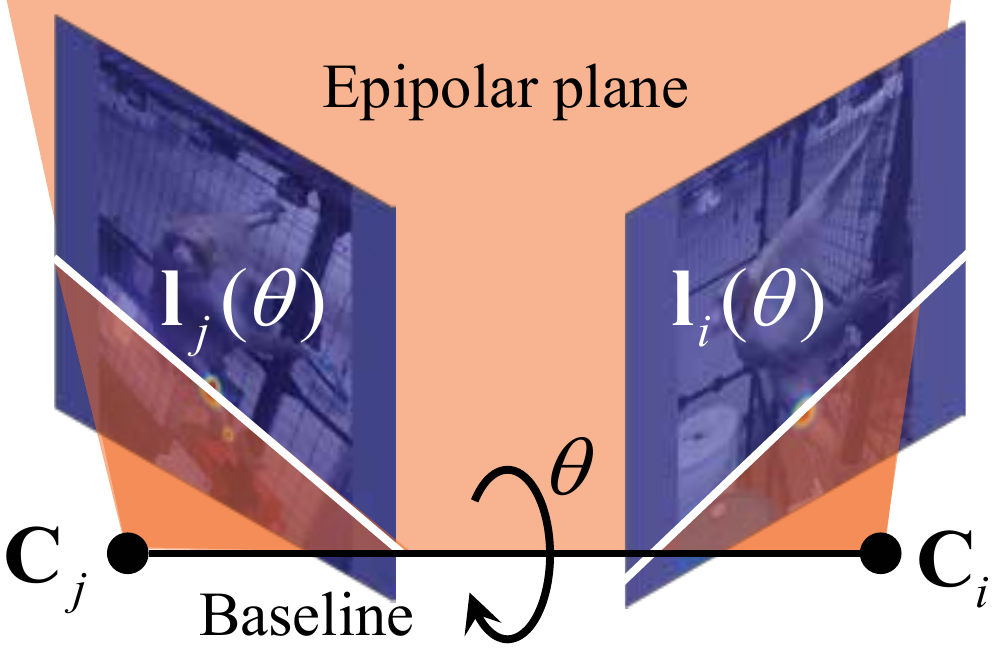}}
\end{center}
\vspace{-3mm}
   \caption{(a) The keypoint distribution of the knee joint for the $j^{\rm th}$ image, $P_j$, is transferred to the $i^{\rm th}$ image to form the epipolar line distribution $P_{j\rightarrow i}(\mathbf{x}_i)$. Note that the points that lie in the same epipolar line have the equal transferred distribution, $P_{j\rightarrow i}(\mathbf{x}_i)=P_{j\rightarrow i}(\mathbf{y}_i)$, and therefore (b) the distribution can be reparametrized by the 1D rotation $\theta\in\mathds{S}$ about the baseline where $\mathbf{C}_i$ and $\mathbf{C}_j$ are the camera optical centers. We match two distributions: the distribution transferred from the $i^{\rm th}$ image $Q_{j\rightarrow i}(\theta)$ and the distribution of keypoint in the $j^{\rm th}$ image $Q_i(\theta)$. The minimization of the epipolar divergence $D_E(Q_i||Q_{j\rightarrow i})$ is provably equivalent to reprojection error minimization. }
\label{Fig:epoch}
\vspace{-4mm}
\end{figure*}



\section{MONET}
We present a semi-supervised learning framework for training a keypoint detector by leveraging multiview image streams for which $|\mathcal{D}_U| \gg |\mathcal{D}_L|$, where $\mathcal{D}_L$ and $\mathcal{D}_U$ are labeled and unlabeled data, respectively.
We learn a network model that takes an input image $\mathcal{I}$ and outputs a keypoint distribution, i.e., $\phi(\mathcal{I};\mathbf{w}) \in [0,1]^{W \times H \times C}$ where $\mathcal{I}$ is an input image, $\mathbf{w}$ is the learned network weights, and $W$, $H$, and $C$ are the width, height, and the number of keypoints.  
To enable end-to-end cross-view supervision without 3D reconstruction, we formulate a novel raster representation of epipolar geometry in Section~\ref{Sec:divergence}, and show how to implement it in practice using stereo rectification in Section~\ref{Sec:cross}. The full learning framework is described in Section~\ref{Sec:overview} by incorporating a bootstrapping prior.


\subsection{Epipolar Divergence} \label{Sec:divergence}

A point in the $i^{\rm th}$ image $\mathbf{x}_i \in \mathds{R}^2$ is {\em transferred} to form a corresponding epipolar line in the $j^{\rm th}$ image via the fundamental matrix $\mathbf{F}$ between two relative camera poses, which measures geometric consistency, i.e., the corresponding point $\mathbf{x}_j$ must lie in the epipolar line~\cite{hartley:2004}:
\begin{align}
    D(\mathbf{x}_i, \mathbf{x}_j) &= \left|\widetilde{\mathbf{x}}_j^\mathsf{T}
    \left(\mathbf{F} \widetilde{\mathbf{x}}_i\right)\right| 
    \propto \underset{\mathbf{x} \in \mathbf{F}\widetilde{\mathbf{x}}_i}{\operatorname{inf}} \|\mathbf{x}-\mathbf{x}_j \|.\label{Eq:epipolar_line}
\end{align}
The infimum operation measures the distance between the closest point in the epipolar line ($\mathbf{F} 
\widetilde{\mathbf{x}}_i$) and $\mathbf{x}_j$ in the $j^{\rm th}$ image. 

We generalize the epipolar line transfer to define the distance between keypoint distributions. Let $P_{i}:\mathds{R}^2\rightarrow [0,1]$ be the keypoint distribution given the $i^{\rm th}$ image computed by a keypoint detector, i.e., $P_i(\mathbf{x}) = \left. \phi(\mathcal{I}_i;\mathbf{w})\right|_{\mathbf{x}}$, and $P_{j\rightarrow i}:\mathds{R}^2\rightarrow [0,1]$ be the keypoint distribution in the $i^{\rm th}$ image {\em transferred} from the $j^{\rm th}$ image as shown in Figure~\ref{Fig:geom1}. 
Note that we abuse notation by omitting the keypoint index, as each keypoint is considered independently. 

Consider a max-pooling operation along a line, $g$:
\begin{align}
    g(\mathbf{l}; P) = \underset{\mathbf{x} \in \mathbf{l}}{\operatorname{sup}}~P(\mathbf{x}), \label{Eq:max_pool}
\end{align}
where $P:\mathds{R}^2\rightarrow [0,1]$ is a distribution and $\mathbf{l} \in \mathds{P}^2$ is a 2D line parameter. $g$ takes the maximum value along the line in $P$. Given the keypoint distribution in the $j^{\rm th}$ image $P_j$, the transferred keypoint distribution can be obtained:
\begin{align}
    P_{j\rightarrow i}(\mathbf{x}_i) = g(\mathbf{F}\widetilde{\mathbf{x}}_i; P_j).  \label{Eq:epipolar_transfer1} 
\end{align}
The supremum operation is equivalent to the infimum operation in Equation~(\ref{Eq:epipolar_line}), where it finds the most likely (closest) correspondences along the epipolar line. The first two images in Figure~\ref{Fig:geom1} illustrate the keypoint distribution transfer via Equation~(\ref{Eq:epipolar_transfer1}). The keypoint distribution in the $i^{\rm th}$ image is deterministically transformed to the rasterized epipolar line distribution in the $j^{\rm th}$ image, i.e., no explicit 3D reconstruction (triangulation or depth prediction) is needed. In fact, the transferred distribution is a posterior distribution of a 3D keypoint given a uniform depth prior. 

$P_i$ and $P_{j\rightarrow i}$ cannot be directly matched because $P_i$ is a point distribution while $P_{j\rightarrow i}$ is a line distribution. A key observation is that points that lie on the same epipolar line in $P_{j \rightarrow i}$ have the same probability, i.e., $P_{i\rightarrow j}(\mathbf{x}_j) = P_{i\rightarrow j}(\mathbf{y}_j)$ if $\mathbf{F}^\mathsf{T} \widetilde{\mathbf{x}}_j \propto \mathbf{F}^\mathsf{T} \widetilde{\mathbf{y}}_j$ as shown in the second image of Figure~\ref{Fig:geom1}. This indicates that the transferred distribution can be 
parametrized by the slope of an epipolar line, $\theta \in \mathds{S}$, i.e., 
\begin{align}
    Q_{j\rightarrow i}(\theta) = g(\mathbf{l}_i(\theta); P_{j\rightarrow i}), \label{Eq:flat_ji}
\end{align}
where $\mathbf{l}_i(\theta)$ is the line passing through the epipole parametrized  by $\theta$ in the $i^{\rm th}$ image, and $Q_{j\rightarrow i}:\mathds{S}\rightarrow[0,1]$ is a flattened 1D distribution across the line. Similarly, the flattened keypoint distribution of $P_i$ can be defined as $Q_{i}(\theta) = g(\mathbf{l}_i(\theta); P_{i})$. 

\begin{thm} Two keypoint distributions $P_i$ and $P_j$ are geometrically consistent, i.e., zero reprojection error, if $Q_i(\theta) = Q_{j\rightarrow i}(\theta)$. \label{Thm:consistency}
\end{thm}
See the proof in Appendix. Theorem~\ref{Thm:consistency} states the necessary condition of zero reprojection: the detected keypoints across views must lie in the same epipolar plane in 3D. Figure~\ref{Fig:geom2} illustrates the epipolar plane that is constructed by the baseline and the 3D ray (inverse projection) of the detected keypoint. Matching $Q_{i}$ and $Q_{j\rightarrow i}$ is equivalent to matching the probabilities of epipolar 3D planes, which can be parametrized by their surface normal ($\theta$).





To match their distributions, we define an {\em epipolar divergence} that measures the difference between two keypoint distributions using relative entropy inspired by Kullback–Leibler (KL) divergence~\cite{Kullback:1951}:
\begin{align}
    D_{\rm E}(Q_i||Q_{j\rightarrow i}) = \int_\mathds{S} Q_i(\theta) \log \frac{Q_i(\theta)}{Q_{j \rightarrow i}(\theta)} {\rm d}\theta. \label{Eq:divergence}
\end{align}
This epipolar divergence measure how two keypoint distributions are geometrically consistent.  

\begin{figure*}
\begin{center}
\includegraphics[width=\textwidth]{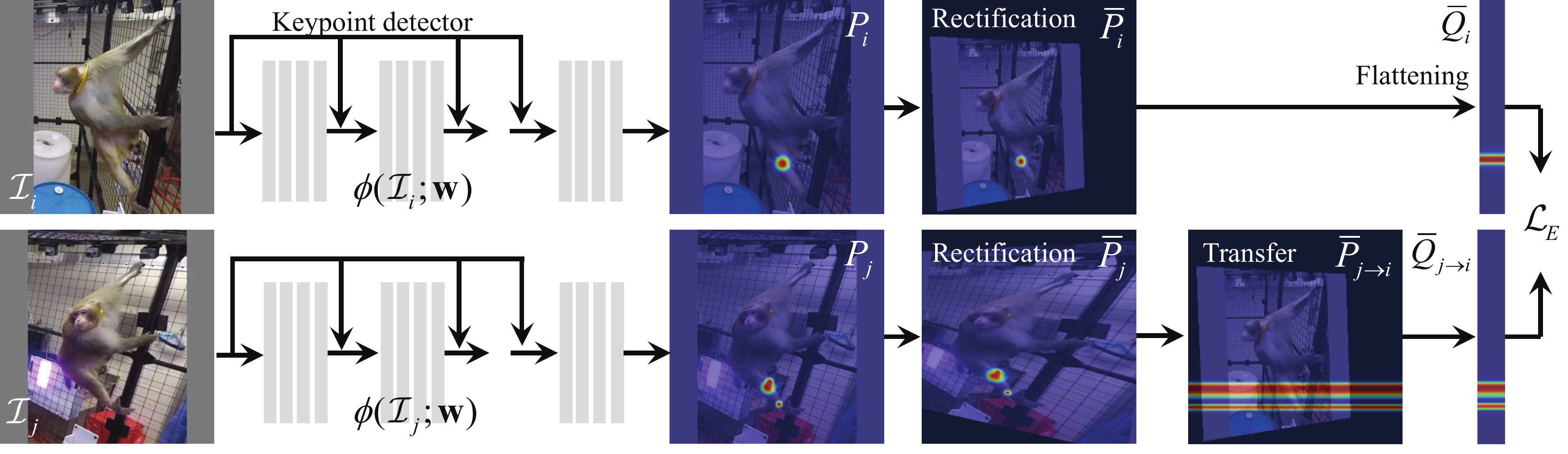}
\end{center}
\vspace{-3mm}
   \caption{We design a twin network to minimize the epipolar divergence between $\overline{Q}_i$ and $\overline{Q}_{j\rightarrow i}$. Stereo rectification is used to simplify the max-pooling operation along the epipolar line, and reduce computational complexity and sampling aliasing.}
\label{Fig:cross}
\vspace{-4mm}
\end{figure*}

\subsection{Cross-view Supervision via Rectification} \label{Sec:cross}

In practice, embedding Equation~(\ref{Eq:divergence}) into an end-to-end neural network is non-trivial because (a) a new max-pooling operation over oblique epipolar lines in Equation~(\ref{Eq:epipolar_transfer1}) needs to be defined; (b) the sampling interval for max-pooling along the line is arbitrary, i.e., uniform sampling does not encode geometric meaning such as depth; and (c) the sampling interval across $\theta$ is also arbitrary. These factors increase computational complexity and sampling artifacts in the process of training. 


We introduce a new operation inspired by stereo rectification, which warps a keypoint distribution such that the epipolar lines become parallel (horizontal) as shown the bottom right image in Figure~\ref{Fig:cross}. This rectification allows converting the max-pooling operation over an oblique epipolar line into regular row-wise max-pooling, i.e., epipolar line can be parametrized by its height $\mathbf{l}(v)$. Equation~(\ref{Eq:max_pool}) can be re-written with the rectified keypoint distribution:
\begin{equation} 
\begin{split}
    \overline{g}(v;\overline{P}) = g\left(\mathbf{l}(v); \overline{P}\right) = \underset{u}{\operatorname{max}}~\overline{P}\left(\begin{bmatrix}u\\v\end{bmatrix}\right) \label{Eq:max_pool1}
\end{split}
\end{equation}
where $(u,v)$ is the $x,y$-coordinate of a point in the rectified keypoint distribution $\overline{P}$ warped from $P$, i.e., $\overline{P}(\mathbf{x}) = P(\mathbf{H}_r^{-1} \mathbf{x})$ where $\mathbf{H}_r$ is the homography of stereo-rectification. $\overline{P}$ is computed by inverse homography warping with bilinear interpolation~\cite{jaderberg:2015,he:2017}.
This rectification simplifies the flattening operation in Equation~(\ref{Eq:flat_ji}):
\begin{equation} 
\begin{split}
    &\overline{Q}_{j\rightarrow i}(v) = \overline{g}(v; \overline{P}_{j\rightarrow i})
    = \overline{g}\left(av+b; \overline{P}_j\right), \\
    &\overline{Q}_{i}(v) = \overline{g}(v; \overline{P}_i), \label{Eq:simple_i}
\end{split}
\end{equation}
where $a$ and $b$ are re-scaling factors between the $i^{\rm th}$ and $j^{\rm th}$ cameras, accounting different camera intrinsic and cropping parameters. See Appendix for more details.



\begin{figure*}
\begin{center}
  \includegraphics[width=1\textwidth]{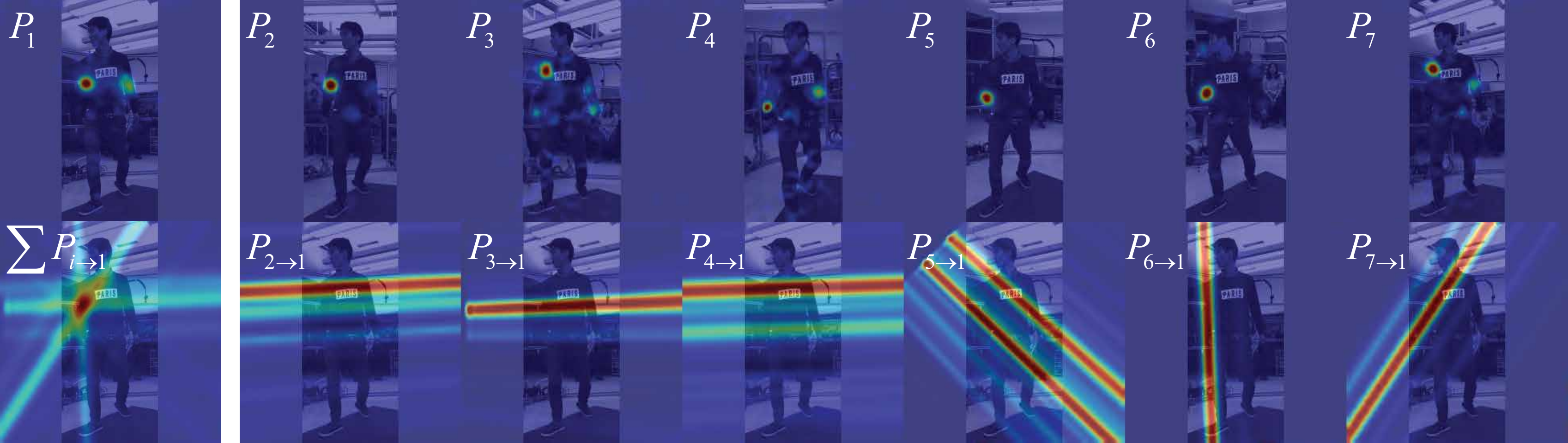}
\end{center}
\vspace{-3mm}
   \caption{Epipolar cross-view supervision on right elbow on view 1. Top right row shows elbow detections across views, i.e., $P_2,\cdots,P_7$. The transferred distribution to view 1 is shown on the bottom right row, i.e., $P_{2\rightarrow 1},\cdots,P_{7\rightarrow 1}$. These transferred probabilities are used to supervise view 1 where the bottom left image is the summation of cross-view supervisions.}
\label{Fig:transfer}
\vspace{-4mm}
\end{figure*}


The key innovation of Equation~(\ref{Eq:simple_i}) is that $\overline{Q}_{j\rightarrow i}(v)$ is no longer 
parametrized by $\theta$ where an additional sampling over $\theta$ is not necessary. It directly accesses $\overline{P}_j$ to max-pool over each row, which significantly alleviates computational complexity and sampling artifacts. Moreover, sampling over the $x$-coordinate is geometrically meaningful, i.e., uniform sampling is equivalent to disparity, or inverse depth. 

With rectification, we model the loss for multiview cross-view supervision:
\begin{align}
    \mathcal{L}_E = \sum_{c=1}^C \sum_{i=1}^S \sum_{j\in \mathcal{V}_i} \sum_{v=1}^H \overline{Q}^c_i(v) \log \frac{\overline{Q}_i^c(v)}{\overline{Q}_{j\rightarrow i}^c(v)}
\end{align}
where $H$ is the height of the distribution, $P$ is the number of keypoints, $S$ is the number of cameras, and $\mathcal{V}_i$ is the set of paired camera indices of the $i^{\rm th}$ camera. We use the superscript in $\overline{Q}_i^c$ to indicate the keypoint index. Figure~\ref{Fig:cross} illustrates our twin network that minimizes the epipolar divergence by applying stereo rectification, epipolar transfer, and flattening operations, which can perform cross-view supervision from unlabeled data.

Since the epipolar divergence flattens the keypoint distribution, cross-supervision from one image can constrain in one direction. In practice, we find a set of images given the $i^{\rm th}$ image 
such that the expected epipolar lines are not parallel. When camera centers lie on a co-planar surface, a 3D point on the surface produces all same epipolar lines, which is a degenerate case\footnote{This degenerate case does not apply for 3D point triangulation where the correspondence is known.}.  Figure~\ref{Fig:transfer} illustrates cross-view supervision on a right elbow on view 1. Elbow detections from view 2 to 7 (top right row) are transferred to view 1 (bottom right row). These transferred probabilities are used to supervise view 1 where the bottom left image is the summation of cross-view supervisions.


\subsection{Multiview Semi-supervised Learning} \label{Sec:overview}

We integrate the raster formulation of the epipolar geometry in Section~\ref{Sec:cross} into a semi-supervised learning framework. The keypoint detector is trained by minimizing the following loss:
\begin{align}
    \underset{\mathbf{w}}{\operatorname{minimize}}~~\mathcal{L}_L + \lambda_e \mathcal{L}_E + \lambda_p \mathcal{L}_B, \label{Eq:final}
\end{align}
where $\mathcal{L}_L$, $\mathcal{L}_E$, and $\mathcal{L}_B$ are the losses for labeled supervision, multiview cross-view supervision, and bootstrapping prior, and $\lambda_e$ and $\lambda_p$ are the weights that control their importance.

Given a set of labeled data ($<$1\%), we compute the labeled loss as follows:
\begin{align}
    \mathcal{L}_L = \sum_{i\in\mathcal{D}_L} \left\|\phi\left(\mathcal{I}_i; \mathbf{w}\right) - \mathbf{z}_i\right\|^2 \label{Eq:label_loss}
\end{align}
where $\mathbf{z}\in [0,1]^{W \times H \times C}$ is the labeled likelihood of keypoints approximated by convolving the keypoint location with a Gaussian kernel.

To improve performance, we incorporate with offline spatiotemporal label augmentation by reconstructing 3D keypoint trajectories using the multiview labeled data inspired by the multiview bootstrapping~\cite{simon:2017}. Given synchronized labeled images, we triangulate each 3D keypoint $\mathbf{X}$ using the camera projection matrices and the 2D labeled keypoints. The 3D reconstructed keypoint is projected onto the rest synchronized unlabeled images, which automatically produces their labels. 3D tracking~\cite{joo_cvpr_2014,Yoon:2018} further increases the labeled data. For each keypoint $\mathbf{X}_t$ at the $t$ time instant, we project the point onto the visible set of cameras. The projected point is tracked in 2D using optical flow and triangulated with RANSAC~\cite{fischler:1981} to form $\mathbf{X}_{t+1}$. We compute the visibility of the point to reduce tracking drift using motion and appearance cues: (1) optical flow from its consecutive image is compared to the projected 3D motion vector to measure motion consistency; and (2) visual appearance is matched by learning a linear correlation filter~\cite{boddeti:2013} on PCA HOG~\cite{dalal:2005}, which can reliably track longer than 100 frames forward and backward. We use this spatiotemporal data augmentation to define the bootstrapping loss:
\begin{align}
    \mathcal{L}_B = \sum_{i\in \mathcal{D}_U} \|\phi(\mathcal{I}_i;\mathbf{w}) - \widehat{\mathbf{z}}_i\|^2.
\end{align}
where $\widehat{\mathbf{z}}\in [0,1]^{W \times H \times C}$ is the augmented labeled likelihood using bootstrapping approximated by convolving the keypoint location with a Gaussian kernel.

\section{Result} \label{Sec:result}
We build a keypoint detector for each species without a pre-trained model, using the CPM network (5 stages). The code can be found in \url{https://github.com/MONET2018/MONET}. To highlight the model flexibility, we include implementations with two state-of-the-art pose detectors (CPM~\cite{cao2017realtime} and Hourglass~\cite{newell:2016}). $\lambda_e=5$ and $\lambda_p = 1$ are used. Our detection network takes an input image (368$\times$368), and outputs a distribution (46$\times$46$\times C$). In training, we use batch size 30, learning rate $10^{-4}$, and learning decay rate 0.9 with 500 steps. We use the ADAM optimizer of TensorFlow with single nVidia GTX 1080. 

\noindent\textbf{Datasets} We validate our MONET framework on multiple sequences of diverse subjects including humans, dogs, and monkeys. (1) \textbf{Monkey subject} 35 GoPro HD cameras running at 60 fps are installed in a large cage ($9'\times12'\times9'$) that allows the free-ranging behaviors of monkeys. There are diverse monkey activities include grooming, hanging, and walking. The camera produces $1280\times 960$ images. 12 keypoints of monkey's pose in 85 images out of 63,000 images are manually annotated. (2) \textbf{Dog subjects} Multi-camera system composed of 69 synchronized HD cameras (1024$\times$1280 at 30 fps) are used to capture the behaviors of multiple breeds of dogs including Dalmatian and Golden Retrievers. Less than 1\% of data are manually labeled. (3) \textbf{Human subject I} A multiview behavioral imaging system composed of 69 synchronized HD cameras capture human activities at 30 fps with 1024$\times$1280 resolution. 30 images out of 20,700 images are manually annotated. This dataset includes a diverse human activities such as dancing, jumping, and sitting. We use a pre-trained CPM model~\cite{cao2017realtime} to generate the ground truth data. (4) \textbf{Human subject II} We test our approach on two publicly available datasets for human subjects: Panoptic Studio dataset~\cite{joo_iccv_2015} and Human3.6M~\cite{h36m_pami}. For the Panoptic Studio dataset, we use 31 HD videos ($1920\times1080$ at 30 Hz). The scenes includes diverse subjects with social interactions that introduce severe social occlusion. The Human3.6M dataset is captured by 4 HD cameras that includes variety of single actor activities, e.g., sitting, running, and eating/drinking.

\setlength{\columnsep}{8pt}
\begin{wrapfigure}{l}{0.19\textwidth}
\hspace{-.25in}
\vspace{-5mm}
  \begin{center}
    \includegraphics[width=0.19\textwidth]{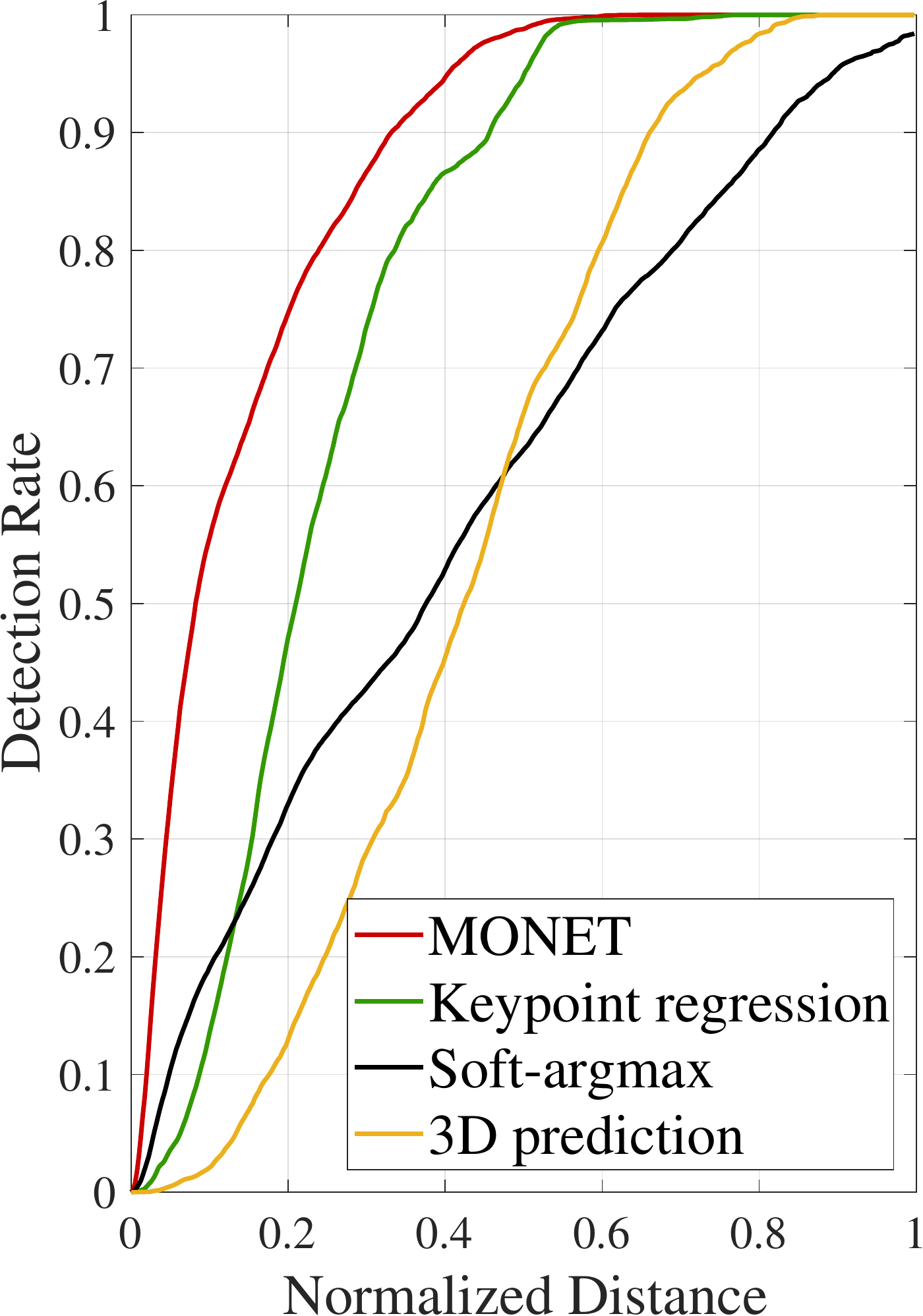}
  \end{center}
    \vspace{-3mm}
  \caption{PCK for hypothesis validation}
  \label{Fig:hypothesis}
  \vspace{-6mm}
\end{wrapfigure}
\noindent\textbf{Hypothesis Validation} We hypothesize that our raster formulation is superior to existing multiview cross-view supervision approaches used for semi-supervised learning because it is an end-to-end system without requiring 3D prediction. We empirically validate our hypothesis by comparing to three approaches on multiview monkey data from 35 views (300 labeled and 600 unlabeled time instances). No pretrained model is used for the evaluation. (1) \textit{Keypoint regression}: a vector representation of keypoint locations is directly regressed from an image. We use DeepPose~\cite{toshev:2014} to detect keypoints and use the fundamental matrix to measure the distance (loss) between the epipolar line and the detected points, $|\widetilde{\mathbf{x}}_2^\mathsf{T}\mathbf{F}\widetilde{\mathbf{x}}_1|$, for the unlabeled data. (2) \textit{Soft-argmax}: a vector representation can be approximated by the raster keypoint distribution using a soft-argmax operation: $\mathbf{x}_{\rm softmax} = \sum_\mathbf{x} P(\mathbf{x})\mathbf{x} / \sum_\mathbf{x} P(\mathbf{x})$, which is
reasonable when the predicted probability is nearly unimodal. This is differentiable, and therefore end-to-end training is possible. However, its approximation holds when the predicted distribution is unimodal. We use CPM~\cite{wei2016cpm} to build a semi-supervised network with epipolar distance as a loss. (3) \textit{3D prediction}: each 3D coordinate is predicted from a single view image where the projection of the 3D prediction is used as cross-view superivison~\cite{Rhodin:2018cvpr,Kanazawa:2018,NIPS2016_6206}. We augment 3D prediction layers on CPM to regress the depth of keypoints~\cite{Pavlakos:2018}. The reprojection error is used for the loss. Figure~\ref{Fig:hypothesis} illustrates the probability of correct keypoint (PCK) curve, showing that our approach using raster epipolar geometry significantly outperforms other approaches.

\begin{figure*}[t]
\begin{center}
\subfigure[Human subject PCK]{\label{Fig:human_pck}\includegraphics[height=0.2\textheight]{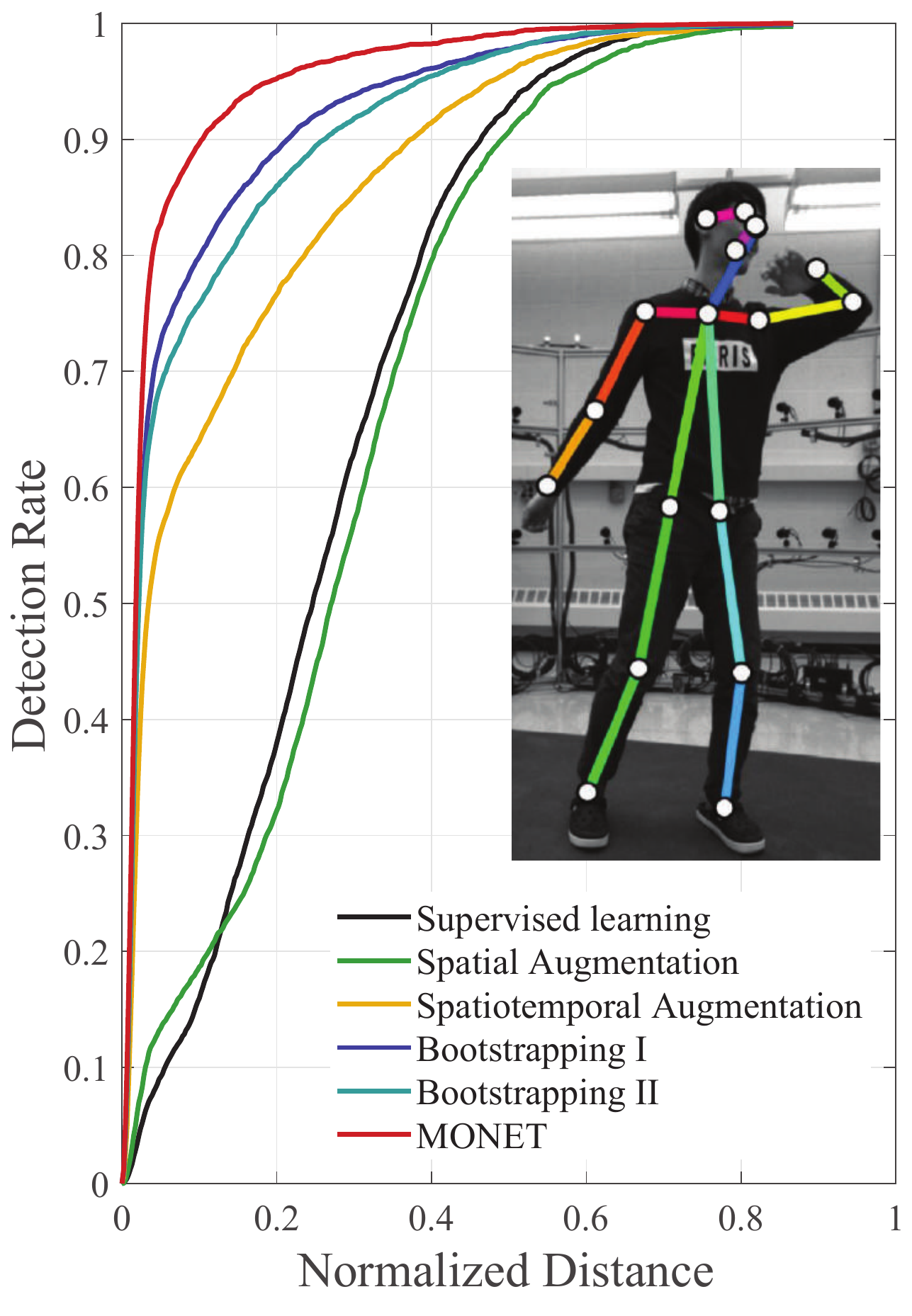}}~~
      \subfigure[Monkey subject PCK]{\label{Fig:monkey_pck}\includegraphics[height=0.2\textheight]{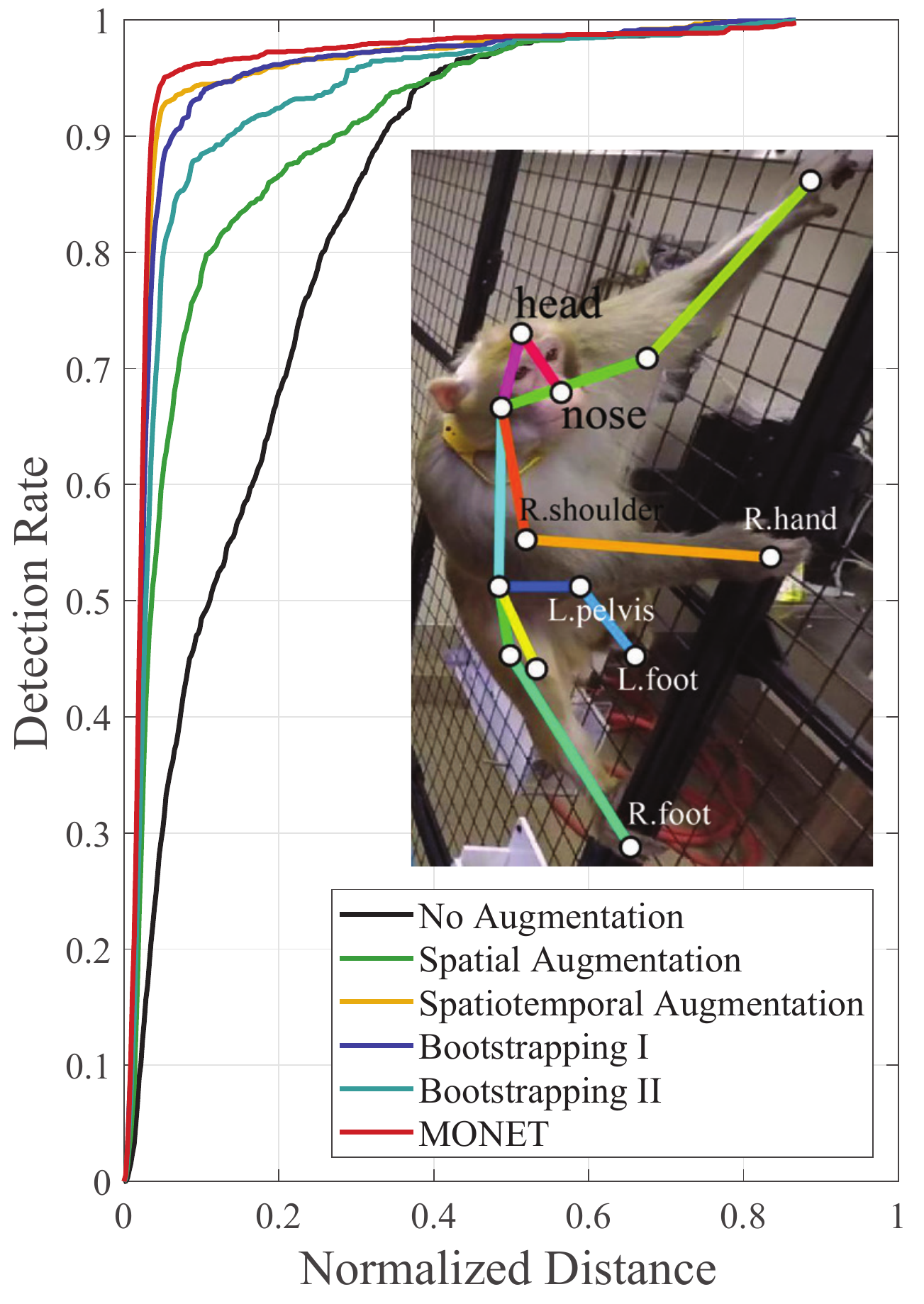}}~~
      \subfigure[Dog subject PCK]{\label{Fig:monkey_pck}\includegraphics[height=0.2\textheight]{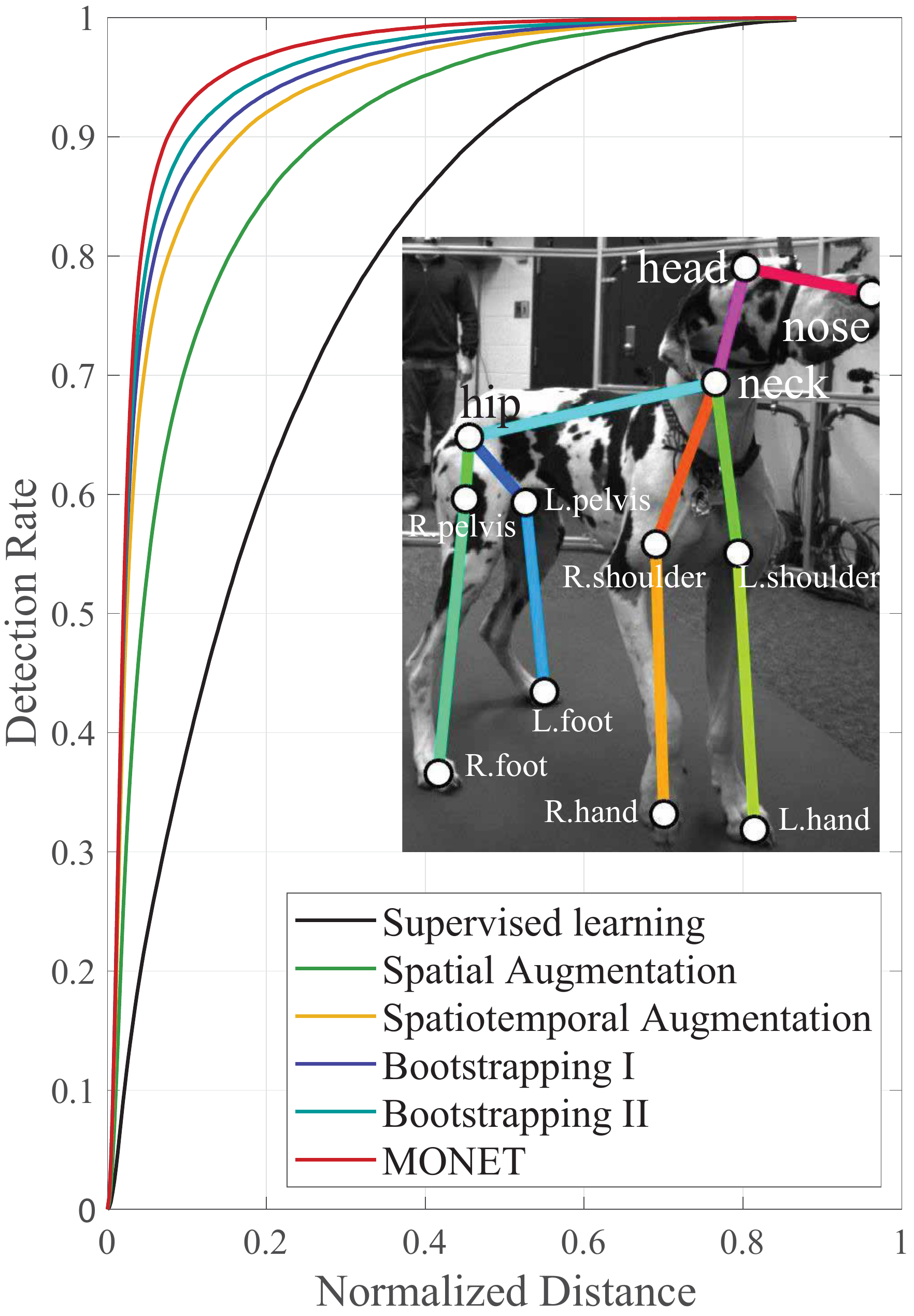}}~~
       \subfigure[Panoptic 
       PCK]{\label{Fig:panoptic_pck}\includegraphics[height=0.2\textheight]{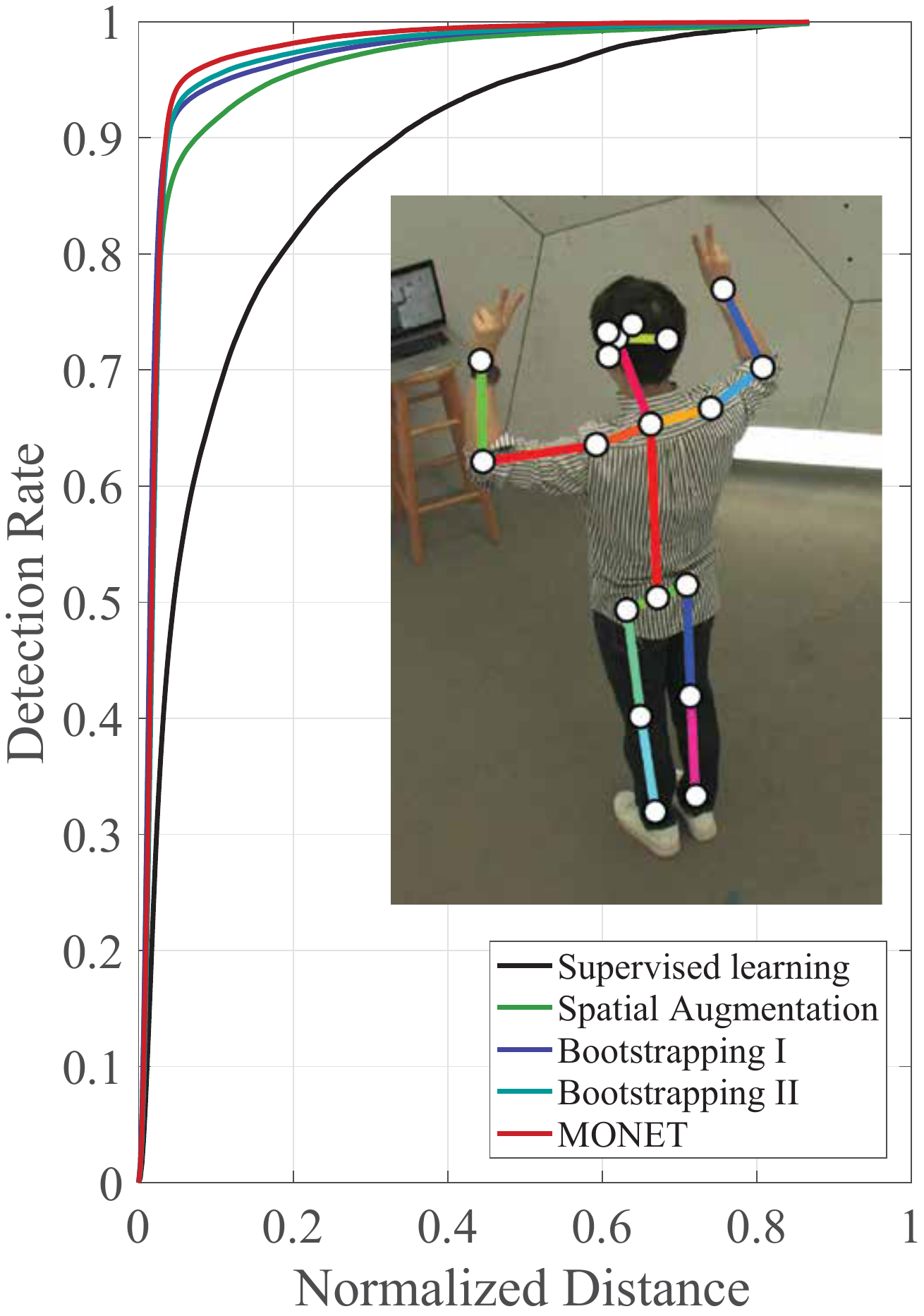}} ~~
      \subfigure[Reprojection error]{\label{Fig:rep}\includegraphics[height=0.2\textheight]{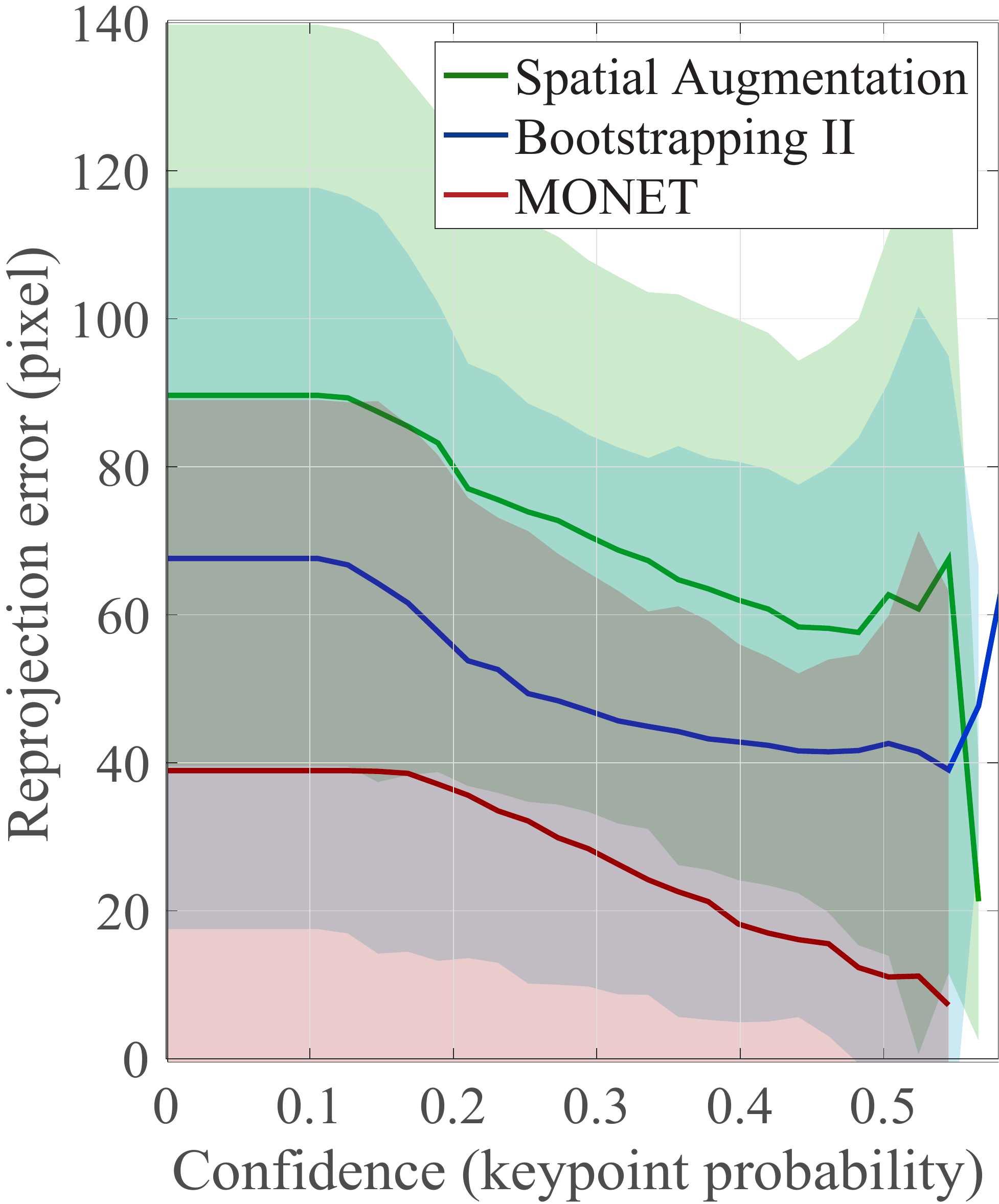}}
     
\end{center}
\vspace{-3mm}
   \caption{PCK curves for (a) humans, (b) monkeys, (c) dogs and (d) the CMU Panoptic dataset~\cite{Panoptic}. MONET (red) outperforms 5 baseline algorithms. (e) MONET is designed to minimize the reprojection error, and we achieve far stronger performance as the confidence increases. }
   \vspace{-4mm}
\label{Fig:pck}
\end{figure*}
\begin{figure*}
  \begin{center}
    \includegraphics[width=1\textwidth]{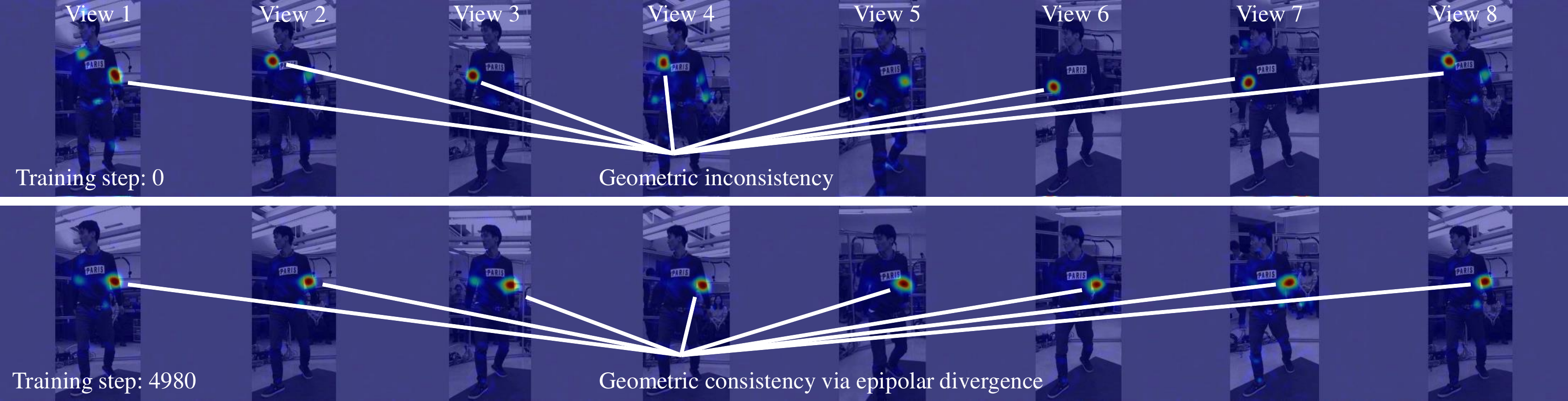}
  \end{center}
    \vspace{-3mm}
  \caption{Erroneous elbow detections from multiview images converge to the geometrically consistent location through training. }
  \label{Fig:elbowconsis}
\vspace{-4mm}
\end{figure*}


\noindent\textbf{Baselines} We compare our approach with 5 different baseline algorithms. For all algorithms, we evaluate the performance on the unlabeled data. (1) \textit{Supervised learning}: we use the manually annotated images to train the network in a fully supervised manner. Due to the limited number of labeled images ($<$100), the existing distillation methods~\cite{hinton:2015, Radosavovic:2017} perform similarly. (2) \textit{Spatial augmentation}: the 3D keypoints are triangulated and projected onto the synchronized unlabeled images. This models visual appearance and spatial configuration from multiple perspectives, which can greatly improve the generalization power of keypoint detection. (3) \textit{Spatiotemporal augmentation}: we track the 3D keypoints over time using multiview optical flow~\cite{joo_cvpr_2014,Yoon:2018}. This augmentation can model different geometric configurations of 3D keypoints. (4) \textit{Bootstrapping I}: Given the spatiotemporal data augmentation, we apply the multiview bootstrapping approach~\cite{simon:2017} to obtain pseudo-labels computed by RANSAC-based 3D triangulation for the unlabeled data. (5) \textit{Bootstrapping II}: the Bootstrapping I model is refined by re-triangulation and re-training. This can reduce the reprojection errors. We evaluate our approach based on accuracy and precision: accuracy measures distance from the ground truth keypoint and precision measures the coherence of keypoint detections across views. (6) \textit{Rhodin et al.~\cite{Rhodin:2018cvpr}}: The unlabeled multi-view image pairs are used to generate 3D point cloud of body first during unsupervised training, and then the model is trained with images with 3D ground truth to learn to transfer point cloud to joint positions.


\noindent\textbf{Accuracy} We use PCK curves to measure the accuracy. The distance between the ground truth keypoint and the detected keypoint is normalized by the size of the width of the detection window (46). Figure~\ref{Fig:pck} shows PCK performance on human, monkey, and dog subjects where no pre-trained model is used. Our MONET (red) model exhibits accurate detection for all keypoints, and outperforms 5 baselines. For the monkey data, higher frame-rate image streams (60 fps) greatly boost the performance of multiview tracking due to smaller displacements, resulting in accurate keypoint detection by spatiotemporal augmentation. We also conducted an experiment on the CMU Panoptic dataset~\cite{Panoptic} to validate the generalization power of our approach. This dataset differs from ours in terms of camera parameters, placements, and scene (e.g., pose, illumination, background, and subject). MONET outperforms on both accuracy (PCK) and precision (reprojection error) as shown in Figure~\ref{Fig:panoptic_pck}. 


\begin{table}
\begin{center}

\small\addtolength{\tabcolsep}{-5pt}

\begin{tabular}{l|c|c|c|c}

\hline
& Human  & Monkey  & Dog  & Panoptic \\
\hline
Supervised learning &77.8$\pm$73.3 &31.1$\pm$872  &88.9$\pm$69.9  &53.2$\pm$271.4\\
Spatial aug. &69.0$\pm$66.2 & 12.9$\pm$26.6 &37.5$\pm$47.1  &22.2$\pm$40.4\\
Spatiotemporal aug. &50.3$\pm$65.4  & 8.10$\pm$17.8 &24.0$\pm$36.2  &N/A\\
Bootstrapping I~\cite{simon:2017} & 28.5$\pm$44.7 &8.68$\pm$18.9 & 18.9$\pm$31.0  &15.6$\pm$31.7\\
Bootstrapping II~\cite{simon:2017} & 35.4$\pm$62.4 & 9.97$\pm$22.1 &17.1$\pm$29.3 &13.7$\pm$24.6\\
MONET & \textbf{15.0}$\pm$\textbf{24.1}& \textbf{5.45}$\pm$\textbf{11.4} &\textbf{10.3}$\pm$\textbf{18.7} 
&\textbf{12.8}$\pm$\textbf{18.0}\\
\hline
\end{tabular}
\end{center}
\vspace{-4mm}
\caption{Reprojection error (Mean$\pm$Std).}\label{table:reprojection}
\vspace{-6mm}
\end{table}

\begin{figure*}[t]
\centering 
\includegraphics[width=0.9\textwidth]{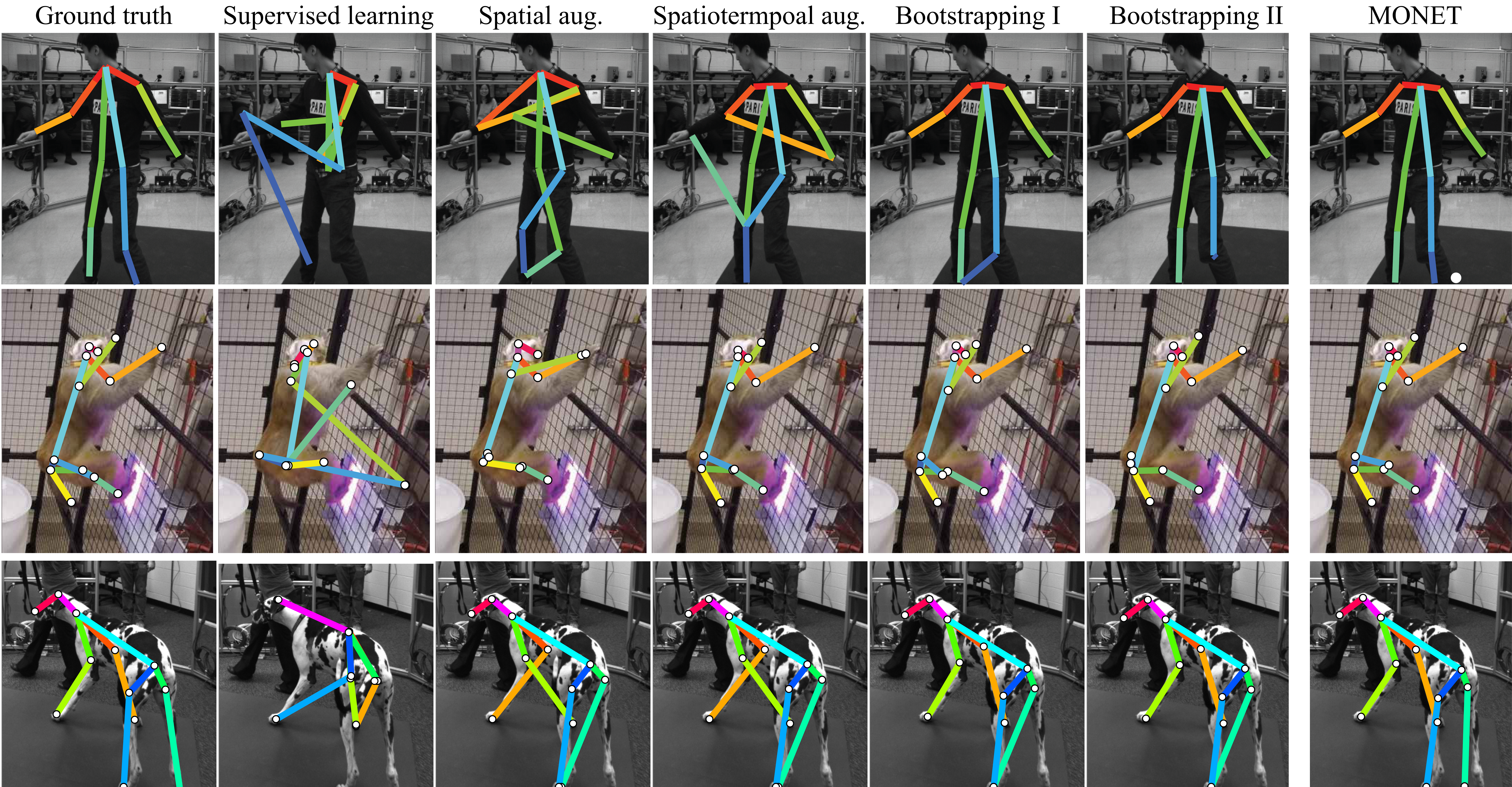}

   \caption{We qualitatively compare our MONET with 5 baseline algorithms on humans, monkeys, and dogs.}
   \vspace{-4mm}
\label{Fig:human_quant}
 \end{figure*}

\noindent\textbf{Precision} We use reprojection error to evaluate the precision of detection. Given a set of keypoint detections in a synchronized frame and 3D camera poses, we triangulate the 3D point without RANSAC. The 3D point is projected back to each camera to compute the reprojection error, which measures geometric consistency across all views. MONET is designed to minimize the reprojection error, and it outperforms baselines significantly in Figure~\ref{Fig:rep}.
Our MONET performs better at higher keypoint distribution, which is key for 3D reconstruction because it indicates which points to triangulate. Figure~\ref{Fig:elbowconsis} shows how erroneous detections of the left elbow from multiview images converge to geometrically consistent elbow locations as the training progresses. The performance for each subject is summarized in Table~\ref{table:reprojection}. 

\begin{table}[t]
\begin{center}
\scriptsize\addtolength{\tabcolsep}{-5pt}
\begin{tabular}{l|c|c|c|c|c|c|c|c|c}
\hline
Labeled / Unlabeled & Hips  &R.Leg &R.Arm &Head  &L.Hand &L.Foot & R.UpLeg & Neck & Total \\
\hline
S1 / S5,6,7,8 &13.0 &3.1 &3.4  &1.0 &6.6 &6.2 &10.9 &1.6&5.5  \\
S1,5 / S6,7,8 &12.7 & 2.2 &2.9  &1.0 &5.2 &\textbf{3.3} & 10.9 &1.6 & 5.2\\
S1,5,6 / S7,8 & \textbf{7.1} & \textbf{2.0} &\textbf{2.7} &\textbf{0.9} &\textbf{5.0} &4.7 &\textbf{5.6} &\textbf{1.5} &\textbf{4.3}\\
\hline
\end{tabular}
\end{center}
\vspace{-4mm}
\caption{Mean pixel error vs. labeled data size on Human3.6M dataset}\label{table:mean}
  \vspace{-6mm}
\end{table}

\begin{figure}[t]
\begin{center}
\subfigure[Monkey subject]{\label{Fig:human_pck}\includegraphics[height=0.17\textheight]{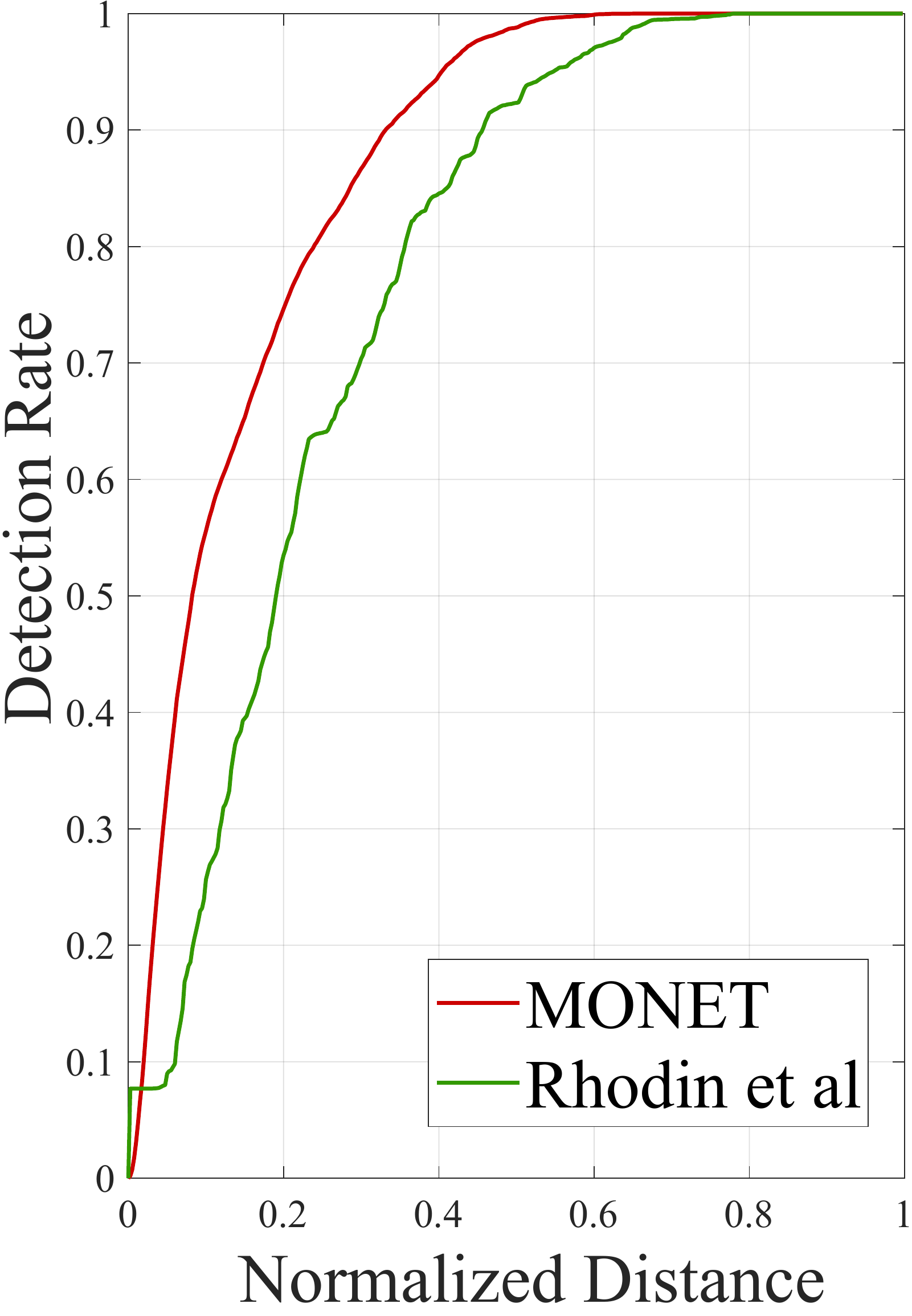}}~~
      \subfigure[Human3.6M]{\label{Fig:monkey_pck}\includegraphics[height=0.17\textheight]{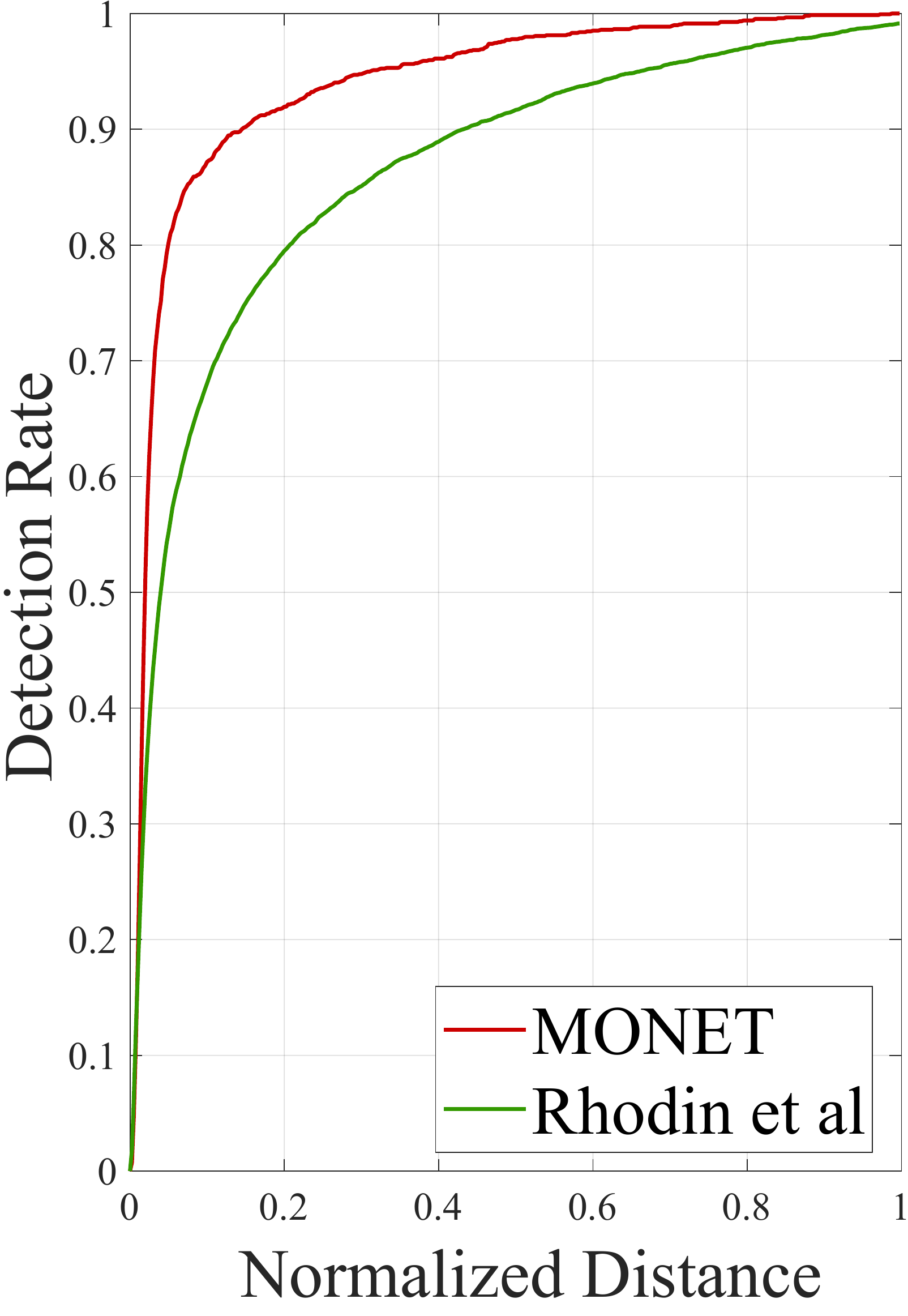}}~~
      \subfigure[Panoptic Studio]{\label{Fig:monkey_pck}\includegraphics[height=0.17\textheight]{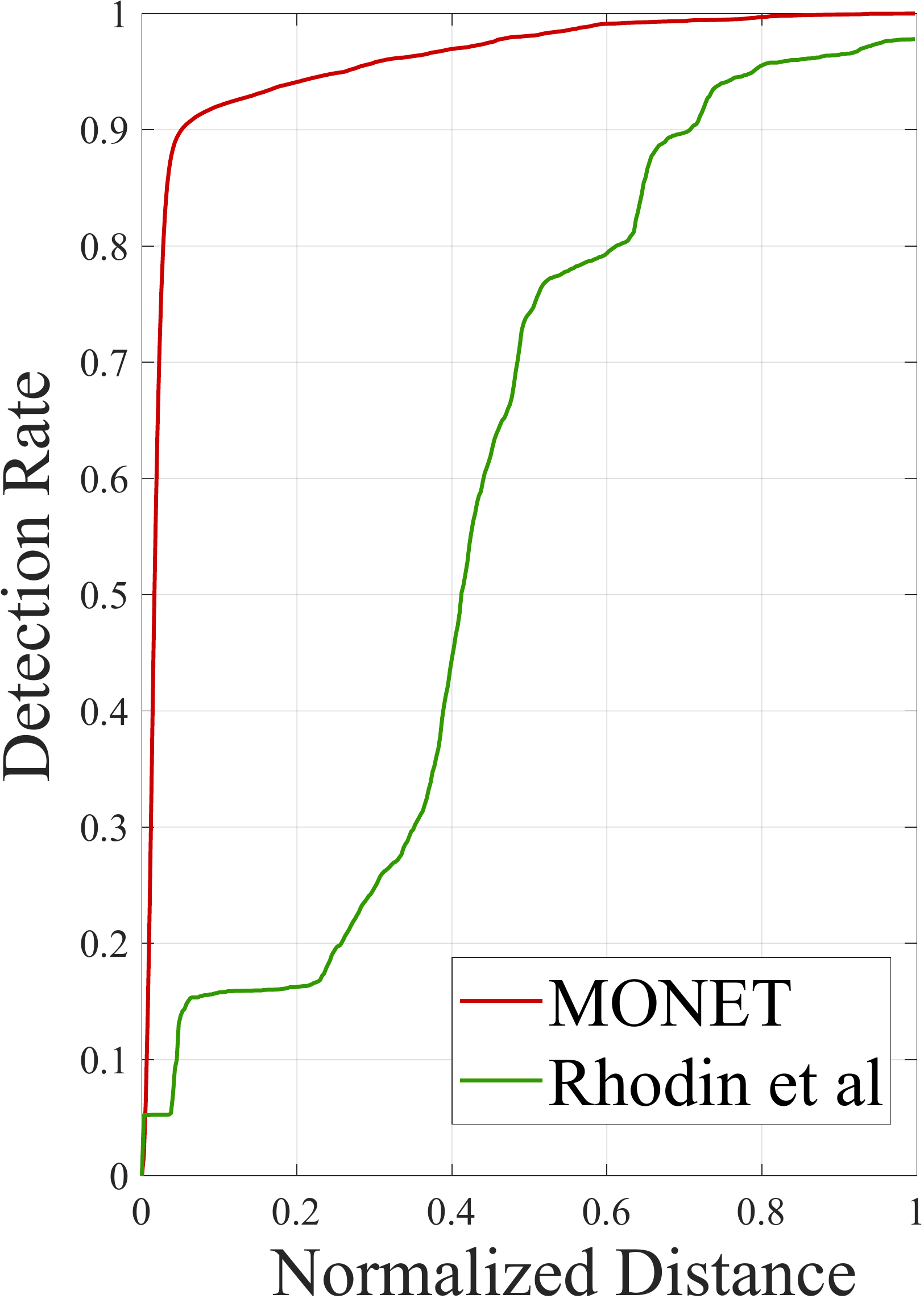}}
\end{center}
\vspace{-3mm}
   \caption{Comparison with Rhodin et al.~\cite{Rhodin:2018cvpr} that predict 3D points for cross-view supervision on monkey, Human3.6M, and Panoptic Studio datasets. }
\label{Fig:pck1}
\vspace{-5mm}
\end{figure}

\noindent\textbf{Robustness} We evaluate the robustness of our approach by varying the amount of labeled data on Human3.6M dataset (four cameras), which provides motion capture ground truth data. Table~\ref{table:mean} summarizes the mean pixel error as varying the labeled and unlabeled subjects. As expected, as the labeled data increases, the error decreases while the minimally labeled S1 (subject 1) still produces less than 15 max pixel error. We also compare to a 3D prediction approach~\cite{Rhodin:2018cvpr}, which showed strong performance on Human3.6M dataset. Similar to their experimental setup, we use S1, S5, and S6 as the labeled data, and S7 and S8 as the unlabeled data for training. In addition to Human3.6M dataset, we also conduct the comparison on the Monkey and CMU Panoptic dataset~\cite{Panoptic}. Figure~\ref{Fig:pck1} illustrates the PCK measure on the unlabeled data. Our approach outperforms the baseline on all the datasets. The advantage of our approach is especially reflected on the CMU Panoptic dataset. Full body is not often visible due to the narrow FOV cameras, which makes the explicit 3D reconstruction in ~\cite{Rhodin:2018cvpr} of body less efficient. 


\noindent\textbf{Qualitative Comparison} A qualitative comparison can be found in Figure~\ref{Fig:human_quant}. MONET can precisely localize keypoints by leveraging multiview images jointly. This becomes more evident when disambiguating symmetric keypoints, e.g., left and right hands, as epipolar divergence penalizes geometric inconsistency (reprojection error). It also shows stronger performance under occlusion (the bottom figure) as the occluded keypoints can be visible to other views that can enforce to the correct location.




\section{Discussion}

We present a new semi-supervised framework, MONET, to train keypoint detection networks by leveraging multiview image streams. The key innovation is a measure of geometric consistency between keypoint distributions called epipolar divergence. Similar to epipolar distance between corresponding points, it allows us to directly compute reprojection error while training a network. We introduce a stereo rectification of the keypoint distribution that simplifies the computational complexity and imposes geometric meaning on constructing 1D distributions. A twin network is used to embed computation of epipolar divergence. We also use multiview image streams to augment the data in space and time, which bootstraps unlabeled data. 
We demonstrate that our framework outperforms existing approaches, e.g., multiview bootstrapping, in terms of accuracy (PCK) and precision (reprojection error), and apply it to non-human species such as dogs and monkeys.
We anticipate that this framework will provide a fundamental basis for enabling {\em flexible} marker-less motion capture that requires exploiting a large (potentially unbounded) number of unlabeled data.

\section{Acknowledgments}

We thank David Crandall for his support and feedback. This work is supported by NSF IIS 1846031.

{\small
\bibliographystyle{ieee_fullname}
\bibliography{egpaper}
}

\clearpage

\appendix

\twocolumn[{%
\begin{center}
\maketitle
\title{\textbf{\Large Supplementary Material}}
\end{center}
}]


\section{Proof of Theorem 1}\label{Sec:proof}

\begin{figure}
  \centering  
    \includegraphics[width=0.35\textwidth]{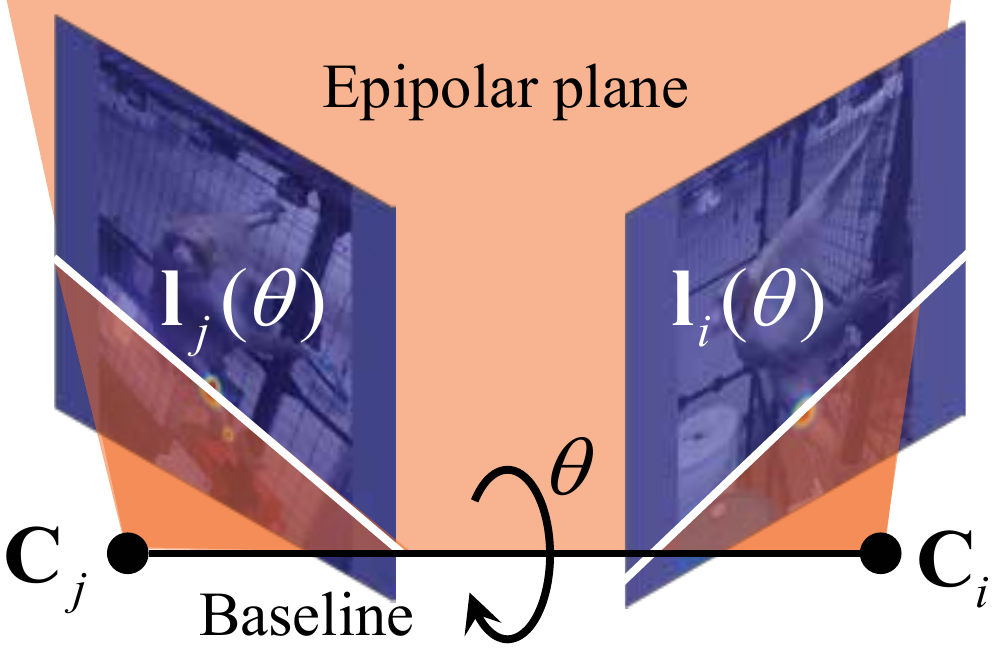}
  \caption{Two epipolar lines are induced by an epipolar plane, which can be parametrized by the rotation $\theta$ about the baseline where $\mathbf{C}_i$ and $\mathbf{C}_j$ are the camera optical centers.} 
  \label{Fig:geom2}
\end{figure}

\begin{proof}
A point in an image corresponds to a 3D ray $\mathbf{L}$ emitted from the camera optical center $\mathbf{C}$ (i.e., inverse projection), and $\lambda$ corresponds to the depth. \textbf{K} is the intrinsic parameter. The geometric consistency, or zero reprojection error, is equivalent to proving $\mathbf{L}^*_i,\mathbf{L}^*_j \in \boldsymbol{\Pi}$ where $\boldsymbol{\Pi}$ is an epipolar plane rotating about the camera baseline $\overline{\mathbf{C}_i \mathbf{C}_j}$ as shown in Figure~\ref{Fig:geom2}, and $\mathbf{L}_i^*$ and $\mathbf{L}_j^*$ are the 3D rays produced by the inverse projection of correspondences $\mathbf{x}_i^*\leftrightarrow\mathbf{x}_j^*$, respectively, i.e., $\mathbf{L}_i^* = \mathbf{C}_i + \lambda \mathbf{R}_i^\mathsf{T} \mathbf{K}^{-1} \widetilde{\mathbf{x}}_i^*$. The correspondence from the keypoint distributions are:
\begin{align}
    \mathbf{x}^*_i &= \underset{\mathbf{x}}{\operatorname{argmax}}~P_i(\mathbf{x})\\
\mathbf{x}^*_j &= \underset{\mathbf{x}}{\operatorname{argmax}}~P_j(\mathbf{x}),
\end{align}


$Q_i(\theta)=Q_{j\rightarrow i}(\theta)$ implies:
\begin{align}
\theta^* &= \underset{\theta}{\operatorname{argmax}}~\underset{\mathbf{x} \in \mathbf{l}_i(\theta)}{\operatorname{sup}}~P_i(\mathbf{x}) \nonumber\\
&= \underset{\theta}{\operatorname{argmax}}~\underset{\mathbf{x} \in \mathbf{l}_i(\theta)}{\operatorname{sup}}P_{j\rightarrow i}(\mathbf{x}) \nonumber\\ 
    &= \underset{\theta}{\operatorname{argmax}}~\underset{\mathbf{x} \in \mathbf{l}_j(\theta)}{\operatorname{sup}}P_{j}(\mathbf{x}).
\end{align}
This indicates the correspondence lies in epipolar lines induced by the same $\theta^*$, i.e,. $\mathbf{x}_i^* \in \mathbf{l}_i(\theta^*)$ and $\mathbf{x}_j^* \in \mathbf{l}_j(\theta^*)$. Since $\mathbf{l}_j(\theta^*) = \mathbf{F}\widetilde{\mathbf{x}}_i^*$, $\mathbf{l}_i(\theta^*)$ and $\mathbf{l}_j(\theta^*)$ are the corresponding epipolar lines. Therefore, they are in the same epipolar plane, and the reprojection error is zero.
\end{proof}

\section{Cropped Image Correction and Stereo Rectification} \label{Sec:crop}
\begin{figure*}
  \centering  
    \includegraphics[width=0.9\textwidth]{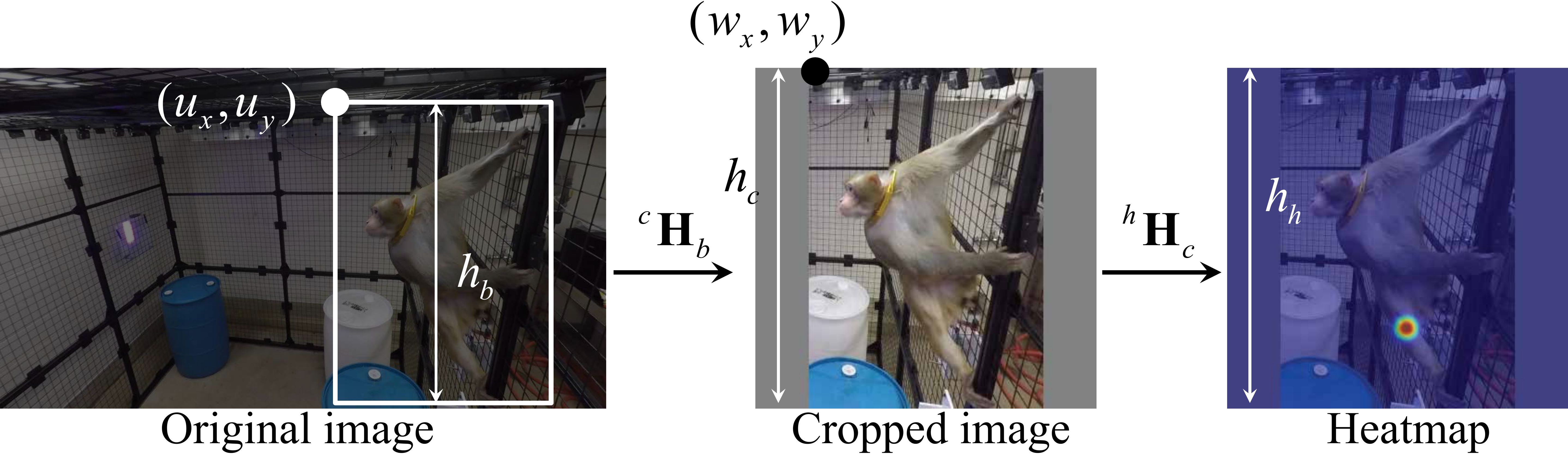}
  \caption{A cropped image is an input to the network where the output is the keypoint distribution. To rectify the keypoint distribution (heatmap), a series of image transformations need to be applied. } 
  \label{Fig:crop}
\end{figure*}
\begin{table*}
\centering
\footnotesize
\small\addtolength{\tabcolsep}{5pt}
\begin{tabular}{l|c|c|c|c|c|c|c}
\hline
Subjects  & $P$ & $|\mathcal{D}_L|$ & $|\mathcal{D}_U|$ & $|\mathcal{D}_L|/|\mathcal{D}_U|$ & $C$ & FPS & Camera type\\
\hline
Monkey & 13 & 85 & 63,000 & 0.13\% & 35 & 60& GoPro 5\\
Humans &14 &30 & 20,700 & 0.14\%& 69& 30& FLIR BlackFly S\\
Dog I  &12 &100 &138,000 & 0.07\% & 69& 30 & FLIR BlackFly S\\
Dog II  &12 &75 &103,500 & 0.07\%& 69& 30 & FLIR BlackFly S\\
Dog III  &12 &80 &110,400& 0.07\% & 69& 30 & FLIR BlackFly S\\
Dog IV  &12 &75 &103,500& 0.07\% & 69& 30 & FLIR BlackFly S\\
\hline
\end{tabular}
\vspace{2mm}
\caption[Summary of multi-camera datasetaaa]{Summary of multi-camera dataset where $P$ is the number of keypoints, $C$ is the number of cameras, $|\mathcal{D}_L|$ is the number of labeled data, and $|\mathcal{D}_U|$ is the number of unlabeled data. }
\label{table:data}
\end{table*}

We warp the keypoint distribution using stereo rectification. This requires a composite of transformations because the rectification is defined in the full original image. The transformation can be written as:
\begin{align}
    ^{\overline{h}}\mathbf{H}_h = \left(^{\overline{h}}\mathbf{H}_{\overline{c}}\right)\left(^{\overline{c}}\mathbf{H}_{\overline{b}}\right)\mathbf{H}_r \left(^c\mathbf{H}_b\right)^{-1} \left(^h\mathbf{H}_c\right)^{-1}.
\end{align}
The sequence of transformations takes a keypoint distribution of the network output $P$ to the rectified keypoint distribution $\overline{P}$: heatmap$\rightarrow$cropped image$\rightarrow$original image$\rightarrow$rectified image$\rightarrow$rectified cropped image$\rightarrow$rectified heatmap. 

Given an image $\mathcal{I}$, we crop the image based on the bounding box as shown in Figure~\ref{Fig:crop}: the left-top corner is $(u_x, u_y)$ and the height is $h_b$. The transformation from the image to the bounding box is:
\begin{align}
    ^c\mathbf{H}_b = \begin{bmatrix}s &  0 & w_x-s u_x\\ 0 & s & w_y-s u_y\\0 & 0 & 1\end{bmatrix}
\end{align}
where $s = h_c/h_b$, and $(w_x,w_y)$ is the offset of the cropped image. It corrects the aspect ratio factor. $h_c=364$ is the height of the cropped image, which is the input to the network. The output resolution (heatmap) is often different from the input, $s_h = h_h/h_c \neq 1$, where $h_h$ is the height of the heatmap. The transformation from the cropped image to the heatmap is:
\begin{align}
    ^h\mathbf{H}_c = \begin{bmatrix}s_h & 0 & 0 \\ 0 & s_h & 0 \\ 0 & 0 & 1\end{bmatrix}
\end{align}
The rectified transformations $\left(^{\overline{h}}\mathbf{H}_{\overline{c}}\right)$ and $\left(^{\overline{c}}\mathbf{H}_{\overline{b}}\right)$ can be defined in a similar way. 

The rectification homography can be computed as $\mathbf{H}_r = \mathbf{K} \mathbf{R}_n \mathbf{R}^\mathsf{T} \mathbf{K}^{-1}$ where $\mathbf{K}$ and $\mathbf{R} \in SO(3)$ are the intrinsic parameter and 3D rotation matrix and $\mathbf{R}_n$ is the rectified rotation of which x-axis is aligned with the epipole, i.e., $\mathbf{r}_x = \dfrac{\mathbf{C}_j - \mathbf{C}_i}{\|\mathbf{C}_j - \mathbf{C}_i\|}$ where $\mathbf{R}_n = \begin{bmatrix}\mathbf{r}_x^\mathsf{T} \\\mathbf{r}_y^\mathsf{T} \\ \mathbf{r}_z^\mathsf{T}\end{bmatrix}$ and other axes can be computed by the Gram-Schmidt process.

The fundamental matrix between two rectified keypoint distributions $\overline{P}_i$ and $\overline{P}_j$ can be written as:
\begin{align}
    \mathbf{F} &= \mathbf{K}_j^{-\mathsf{T}} \begin{bmatrix}1\\0\\0\end{bmatrix}_\times` \mathbf{K}_i^{-1}\nonumber\\
    &= \begin{bmatrix}0 & 0 & 0\\0 & 0 & -1/f_y^j\\0 & 1/f_y^i & p_y^j/f_y^j-p_y^i/f_y^i\end{bmatrix}
\end{align}
where $[\cdot]_\times$ is the skew symmetric representation of cross product, and 
\begin{align}
    \mathbf{K}_i = \begin{bmatrix}f_x^i & 0 & p_x^i\\0 & f_y^i & p_y^i \\ 0 & 0 & 1\end{bmatrix}.
\end{align}
This allows us to derive the re-scaling factor of $a$ and $b$ in Equation~(\ref{Eq:simple_i}):
\begin{align}
    a &= \frac{s^i f_y^i}{s^j f_y^j}\\
    b &= s_h s^i \left(\left(\overline{u}_y^j-p_y^j\right)\frac{f_y^i}{f_y^j}+p_y^i-\overline{u}_y^i\right)
\end{align}
where $\overline{u}_y^i$ is the bounding box offset of the rectified coordinate.

\section{Evaluation Dataset} \label{Sec:data}
All cameras are synchronized and calibrated using structure from motion~\cite{hartley:2004}. The input of most pose detector models except for~\cite{cao2017realtime} is a cropped image containing a subject, which requires specifying a bounding box. We use a kernelized correlation filter~\cite{henriques:2015} to reliably track a bounding box using multiview image streams given initialized 3D bounding box from the labeled data. 

\end{document}